%% file: master_tase.tex
\documentclass[journal]{IEEEtran}
\usepackage[T1]{fontenc}
\usepackage[latin9]{inputenc}
\usepackage{geometry}
\usepackage{float}
\usepackage{amsmath}
\usepackage{amssymb}
\usepackage{graphicx}
\usepackage{esint}
\usepackage[english]{babel}

\include{preamble}

\floatstyle{ruled}
\newfloat{algorithm}{tbp}{loa}
\providecommand{\algorithmname}{Algorithm}
\floatname{algorithm}{\protect\algorithmname}

\begin{document}

\title{Large Scale Estimation in Cyberphysical Systems using Streaming Data:
a Case Study with Smartphone Traces}

\author{Timothy~Hunter,
        Tathagata~Das,
        Matei~Zaharia,
        Pieter~Abbeel,
        and Alexandre~M.~Bayen%

\thanks{T. Hunter, T. Das, M. Zaharia, P. Abbeel and A. Bayen are with the Department
of Electrical and Computer Science, University of California at Berkeley, CA,
94709 USA e-mail: \{tjhunter,tdas,matei\}@eecs.berkeley.edu, pabbeel@cs.berkeley.edu and bayen@berkeley.edu.}

\thanks{Manuscript received XXX 2012.}}









\maketitle

\begin{abstract}
Controlling and analyzing cyberphysical and robotics systems is increasingly 
becoming a Big Data challenge.
Pushing this data to, and processing in the cloud is more efficient than on-board
processing. However, current cloud-based solutions are not suitable for the latency 
requirements of these applications. We present a new concept, \emph{Discretized Streams} 
or D-Streams, that enables massively scalable computations on streaming data with latencies 
as short as a second.
 We experiment with an implementation of D-Streams on top
of the Spark computing framework. We demonstrate the usefulness of this concept with a novel algorithm
to estimate vehicular traffic in urban networks. Our online EM algorithm can estimate traffic on a very large
city network (the San Francisco Bay Area) by processing tens of thousands of observations per second, with a
latency of a few seconds.


\textit{Note to Practitioners} 

This work was driven by the need to estimate vehicular traffic at a large scale,
in an online setting, using commodity hardware. Machine Learning algorithms 
combined with streaming data are not new, but it still requires deep expertise 
both in Machine Learning
and in Computer Systems to achieve large scale computations in a tractable manner.
 The Streaming Spark project aims at providing an interface that abstracts out 
all the technical details of the computation platform (cloud, HPC, workstation, etc.).

As shown in this work, Streaming Spark is suitable for implementing and calibrating
non-trivial algorithms on a large cluster, and provides an intuitive yet powerful
programming interface. The readers are invited to refer to the source code referred
in this article for more examples.

This article presents algorithms to sample and compute densities for Gamma random
variables restricted to a hyperplane (i.e. distributions of the form 
$T_{i}|\sum_{j}\alpha_{j}T_{j}=d$ with $T_j$ independant Gamma distributions). It is
common in this case to use Gaussian random variables because of closed form solutions
to solve. If one considers positive valued distributions with heavy tails, our formulas
using gamma distributions may be more suitable.
\end{abstract}












%



\newcommand\blfootnote[1]{%
  \begingroup
  \renewcommand\thefootnote{}\footnote{#1}%
  \addtocounter{footnote}{-1}%
  \endgroup
}

\IEEEpeerreviewmaketitle

\blfootnote{GIT COMMIT ID = \input{revision} (To remove in final proof.)}

\input{intro.tex}

\input{dstreams.tex}

\input{model.tex}

\input{experiments.tex}

\input{conclusion.tex}

\bibliographystyle{plain}
\bibliography{../biblio_arterial_spark,../mendeley_biblio}



\input{gammadir.tex}


\input{code.tex}

\end{document}

%% file: preamble.tex
\usepackage[noadjust]{cite}
\usepackage{url}
\usepackage{xspace}
\usepackage{arabtex}
\newcommand{\sys}{Spark Streaming\xspace}

\newcommand{\dstreams}{D-Streams\xspace}
\newcommand{\dstream}{D-Stream\xspace}

\newcommand{\mm}{\emph{Mobile Millennium}\xspace}
\newcommand{\SBO}{SlidingBig}
\newcommand{\SBA}{SlidingBig1}
\newcommand{\SBB}{SlidingBig2}
\newcommand{\SBC}{SlidingBig3}
\newcommand{\SBD}{SlidingBig4}

\usepackage{draftwatermark}
\SetWatermarkLightness{1.0}

\usepackage{listings}

\lstdefinelanguage{scala}{
  morekeywords={abstract,case,catch,class,def,%
    do,else,extends,false,final,finally,%
    for,if,implicit,import,match,mixin,%
    new,null,object,override,package,%
    private,protected,requires,return,sealed,%
    super,this,throw,trait,true,try,%
    type,val,var,while,with,yield},
  otherkeywords={=,=>,<-,<\%,<:,>:,\#,@},
  sensitive=true,
  morecomment=[l]{//},
  morecomment=[n]{/*}{*/},
  morestring=[b]",
  morestring=[b]',
  morestring=[b]"""
}

\definecolor{dkgreen}{rgb}{0,0.6,0}
\definecolor{dkblue}{rgb}{0,0,1.0}
\definecolor{gray}{rgb}{0.5,0.5,0.5}
\definecolor{dkgray}{rgb}{0.3,0.3,0.3}
\definecolor{mauve}{rgb}{0.58,0,0.82}
 
\newcommand{\lstfontfamily}{\ttfamily}

\definecolor{darkviolet}{rgb}{0.5,0,0.4}
\definecolor{darkgreen}{rgb}{0,0.4,0.2} 
\definecolor{darkblue}{rgb}{0.1,0.1,0.9}
\definecolor{darkgrey}{rgb}{0.5,0.5,0.5}
\definecolor{lightblue}{rgb}{0.4,0.4,1}

\lstdefinestyle{eclipse}{
	basicstyle=\small\lstfontfamily,
    emphstyle=\color{red}\bfseries, 
    keywordstyle=\color{darkviolet}\bfseries,
    commentstyle=\color{darkgreen},
    stringstyle=\color{darkblue},
    numberstyle=\color{darkgrey}\lstfontfamily,
    emphstyle=\color{red},
    morecomment=[s][\color{lightblue}]{/**}{*/},
  showstringspaces=false,
  numbers=left
}

%% file: revision.tex
44d6af

%% file: intro.tex
\section{Introduction}

\label{sec:intro}

Cyber-physical systems involve a complex integration of physical and 
computational processes. Such systems usually integrate two 
distinct components - (i) a set of sensors that continuously produce streaming data 
(ii) a set of communication and computation systems that aggregate data
and perform data analytics. Large-scale cyber-physical systems can be found 
everywhere - from
intricate control systems used in robotics to complex environment sensing of civil 
infrastructures. The data produced by such systems can be very large (millions of  
records/seconds) and the amount of computation necessary to interpret
 the data can be significant. 

Furthermore, over the last decade, the cost of sensing and communication equipment 
has fallen dramatically to the point that it is less expensive to
incorporate a large number of sensors collecting low-value information,
rather than judiciously deploying a limited number of sophisticated and
more accurate measurement systems. The corresponding fall in costs shifts the burden
from carefully designing instrumentation to correctly \emph{filtering
and interpreting} the wealth of information available to  the researcher.
This paradigm shift leads to two problems: storage and processing.
It is not clear which part of the information is relevant so potentially all the
information needs to be saved, which may be expensive: in order to
find the proverbial needle in the haystack, one needs enough room
for the haystack. In addition, small, cheap, unreliable sensors provide
 information that is more noisy and potentially requires more
processing compared to dedicated sensors.

Industries such as genomics and astronomy have learned to cope
with extremely large datasets over the last decade. What makes cyberphysical 
systems truly stand out amongst these applications
is the \emph{very fast decay of the value of information}: in robotics
systems for example, the data collected is usually fed into a control system.
 Past information is often of limited or no value, sometimes
as fast as in the span of a few minutes or tens of seconds. This is unlike
genomic records which, rather than being processed immediately, need 
to be stored reliably for a long time. In essence, the incoming information in
cyber-physical systems needs to be processed as a stream, and not so much
as an ever-growing dataset. 

The problem of processing large incoming streams of data has received
little attention so far, because most work has focused on adapting existing technologies
which are either design scalability or latency but not both. Streaming databases~\cite{streambase} can provide the necessary low latencies 
but are limited in scalability. On the other hand, scalable batch processing 
systems like MapReduce or Hadoop \cite{hadoop} are designed for scaling to 
thousands of machines but perform poorly in terms of latency. Latency is 
not a concern for many applications, and for these running regular batch jobs with
traditional batch systems is appropriate. 
But this is often insufficient for many cyber-physical systems such 
as robotics. Working at lower latencies (at the scale of seconds or tens of seconds) 
presents significant challenges. As a solution
to this problem, we use \dstreams \cite{zaharia2012discretized},
a recently proposed programming model where streaming computations are 
decomposed into a \emph{series of batch computations on small time intervals}.
This model provides two significant benefits.
\begin{itemize}
\item \textit{Scalability with low latency:} Stream processing applications, 
implemented using \dstreams to scale to large clusters (hundreds of cores) while 
providing latencies as low as hundreds of milliseconds. 
\item \textit{Simple high-level programming API:} The \dstream abstraction and its 
associated operations makes it very convenient for a developer to implement 
complex business logic to process the raw sensor data. 
\end{itemize}

In this article, we investigate the use of \sys, a system implementing \dstreams,
 with a large scale estimation problem: inferring the state of traffic on a large
road network by using streams of GPS readings. Traffic congestion
affects nearly everyone in the world due to the environmental damage
and transportation delays it causes. The~$2007$ Urban Mobility Report~\cite{tti}
states that traffic congestion causes~$4.2$ billion hours of extra
travel in the United States every year, which accounts for 2.9 billion
extra gallons of fuel and an additional cost of~\$78 billion. Providing
drivers with accurate traffic information reduces the stress associated
with congestion and allows drivers to make informed decisions, which
generally increases the efficiency of the entire road network~\cite{ban_optimal_2008}.

Modeling highway traffic conditions has been well-studied by the transportation
community with work dating back to the pioneering work of Lighthill,
Whitham and Richards~\cite{lighthill_kinematic_1955}. Recently,
researchers demonstrated that estimating highway traffic conditions
can be done using only GPS probe vehicle data~\cite{work2010traffic}.
Arterial roads, which are major urban city streets that connect population
centers within and between cities, provide additional challenges for
traffic estimation. Recent studies focusing on estimating real-time
arterial traffic conditions have investigated traffic flow reconstruction
for single intersections using dedicated traffic sensors. Dedicated
traffic sensors are expensive to install, maintain and operate, which
limits the number of sensors that governmental agencies can deploy
on the road network. The lack of sensor coverage across the arterial
network thus motivates the use of GPS probe vehicle data for estimating
traffic conditions.

Recent studies focusing on estimating real-time arterial traffic conditions
have investigated traffic flow reconstruction for single intersections~\cite{skabardonis_2005,ban_delay}
using dedicated traffic sensors. We consider an estimation engine
deployed inside the \mm traffic information system \cite{mm,mm-socc}.
This engine gathers GPS observations from participating vehicles and
produces estimates of the travel times on the road network. \mm is
intended to work at the scale of large metropolitan areas: the road
network considered in this work is a real road network (a large portion
of the greater Bay Area, comprising 506,685 road links) and the data
for this work is collected from thousands of vehicles that generate
millions of observations per day. As a consequence of these specifications and requirements,
we employ highly scalable traffic algorithms.

The specific problem we address in this use case is how to extract
travel time distributions from \emph{sparse, noisy} GPS measurements
collected \emph{in real-time} from vehicles. A probabilistic model
of travel times on the arterial network is presented along with an
online \emph{Expectation Maximization} (EM) algorithm for learning the parameters
of this model (Section \ref{sec:model}). The algorithm is expensive
due to the large dimension of the network and the complexity inherent
to the evolution of traffic. Furthermore, our EM algorithm has no
closed-form expression and requires sampling and non-linear optimization
techniques. This is why the use of a distributed system is appropriate.
We will present \dstreams in more detail in Section \ref{sec:d-streams}.

The present work is novel for several reasons. First, it
advances research in traffic estimation by presenting a novel travel
time estimation algorithm that is highly scalable, uses data commonly
available nowadays, and is robust to noise and other random perturbations.
This algorithm builds upon a novel statistical distribution:
the Gamma-Dirichlet distribution (formally introduced in Appendix A).
 Second, it shows that it is possible
to use complex, multistage filtering algorithms on very large systems
with a latencies under a few seconds. Third, it explores the tradeoffs
between computational power, timeliness, and accuracy of estimation
of the travel times outputs and shows that the estimates gracefully
degrade with less data. Fourth, the workflow of this algorithm is
representative of a large class of Machine Learning and estimation
algorithms. We believe the good system performance results obtained
for this particular application (Section \ref{sec:experiments}) hint
at potentially significant speedups for other distributed estimation
algorithms, and are of interest to researchers using cloud computing
for large physical systems, and for the Machine Learning community.

We start by introducing the \dstreams programming model (Section \ref{sec:d-streams}) and the problem of traffic estimation (Section \ref{sec:model}).
We then give an overview of our estimation algorithm (Section \ref{sub:basic-model}) and we explain how we used \sys to parallelize the algorithm
(Section \ref{sub:mm-pipeline}). We evaluate our implementation in
Section \ref{sec:experiments} from the perspective of scalability
(Section \ref{sub:experiment-scalability}) and accuracy (Section
\ref{sub:experiment-ml}). The derivations related to the properties
of the Gamma-Dirichlet distribution are included in the Appendix A.

%% file: dstreams.tex
\section{Discretized streams: large-scale real-time processing of data streams}
\label{sec:d-streams}
Discretized stream (\dstream) is a recently proposed programming model~\cite{zaharia2012discretized} for 
processing of streaming data that allows complex machine learning algorithms to 
be easily expressed and executed on large streams of live data. In this section, we 
will first discuss the limitations of existing techniques of processing live data. Then 
we will elaborate on \dstreams highlighting its advantages over existing techniques. 

\subsection{Limitations of current techniques}
Current techniques to process large amounts of live streaming data can be broadly 
classified into the following two categories.
\begin{itemize}
\item \textit{Using traditional streaming processing systems}: Streaming databases 
like StreamBase~\cite{streambase} and Telegraph~\cite{telegraph}, and stream processing 
 systems like Storm~\cite{storm} have been 
used to meet such processing requirements. While they do achieve low latencies, 
they either have limited fault-tolerance properties (data lost on machine failure) or 
limited scalability (cannot be run on large clusters).  
\item \textit{Using traditional batch processing systems}: The live data is stored 
reliably in a replicated file system like HDFS~\cite{hadoop} and later processed in large 
batches (minutes to hours) using traditional batch processing frameworks like 
Hadoop~\cite{hadoop}. By design, these systems can process large volumes of data on large 
clusters in a fault-tolerant manner, but they can only achieve latencies of minutes at 
best. Furthermore, the processing model is too low level to conveniently express 
complex stream computations.
\end{itemize}
\subsection{\dstream\ - A programming model for stream processing}
D-Streams execute deterministic computations similar to those in MapReduce for fault tolerance, but they do so at a much lower latency than previous systems, by keeping state in memory. The input data received from various input sources (e.g., webservices, 
sensors, etc) during each interval is stored reliably across the cluster to form an 
input dataset for that interval. Once the time interval completes, this dataset is 
processed via deterministic parallel operations (like map, filter, reduce, groupBy, 
etc) to produce new datasets representing program outputs or intermediate states. 
Finally, these datasets can be saved to external source (databases, etc) or aggregate 
all the values into some gradient update / expected loss estimate. The advantage of 
this model is that it provides the developer a convenient high-level programming 
model to easily express complex stream computations while allowing the 
underlying system to process the data in small batches thus achieving excellent 
fault-tolerance properties.

Going into more details, each dataset created in the time intervals is represented as 
a \emph{Resilient Distributed Dataset} (RDD)~\cite{spark} which is  an efficient storage 
abstraction that keeps a distributed collection of data in memory (as opposed to 
writing it to the disk) to guarantee fast access. A \dstream is therefore a series of 
RDDs and lets the user manipulate them collectively through various deterministic 
parallel operations. We illustrate this abstraction and a few operators with a small 
program (written for \sys, an implementation of \dstreams) that computes a 
running count of page view events by URL.
\texttt{}
\begin{lstlisting}
val pageViews = readStream("http://...", "1s")
val ones = pageViews.map(evt => (evt.url, 1))
val counts = ones.runningReduce((a, b) => a+b)
\end{lstlisting}

This code creates a \dstream called {\small pageViews} by reading
an event stream over HTTP, and divides the streams into batches of 1-second intervals.
It then transforms the event stream using the \texttt{map} operator to get a \dstream of (URL, 1) pairs called \texttt{\small ones}, and performs a running count of these
using the \texttt{runningReduce} operator. The arguments to \texttt{map}
and \texttt{runningReduce} are Scala syntax for a closure (function literal). \\

\noindent\subsubsection*{\dstream operators}
\dstreams provide two types of operators to let users build streaming programs:
\begin{itemize}
\item\textit{Transformation operators}, which produce a new \dstream from one or 
more parent streams. These can be either stateless (i.e., act independently on each 
interval) or stateful (share data across intervals).
\item\textit{Output operators}, which let the program write data to external 
systems (e.g., save processed data to a database).
\end{itemize}
\dstreams support the same stateless transformations available in typical batch 
frameworks, including map, reduce, groupBy, and join. In addition, \dstream also 
provide stateful operators like windowing and moving average operators that share 
data across time intervals. The \emph{runningReduce} operator in the earlier page 
view program is an example of a stateful operator as it combines page views across 
time intervals.

Furthermore, since the \dstream abstraction follow the same processing model as 
batch systems, the two can naturally be combined. For example, one may not only 
join two streams of data, but also join a data stream with a batch data - joining a 
stream of incoming tweets against a pre-computed spam filter or historical data. \\

\noindent\subsubsection*{Fault-tolerance properties} 
All the intermediate data computed 
using \dstream are by design fault-tolerant, that is, no data is lost if any machine fails. 
This is achieved by treating each batch of data as an RDD. Each RDD maintains 
a lineage of operations that was used to create it from the raw input data (stored 
reliably by the system by automatic replication)~\cite{spark}. Hence, in the event of 
failure, if any partition of an RDD is lost, it can be recomputed from raw input data 
using the lineage. As these operations are deterministic and functional by nature 
(i.e. side-effect free), the recomputation can be done using fine-grained tasks in 
parallel on many number. This ensures fast recovery minimizing the effect of the 
failure on the stream processing system. This novel technique is called 
\textit{parallel-recovery} and sets this abstraction apart from existing stream processing systems.

\subsection{\sys  - An implementation of \dstreams}
To implement \dstreams , we use Spark, an existing open-source, batch processing 
framework, to create \sys. Spark is a fast, in-memory batch processing framework 
built on the RDD abstraction, and we naturally extend this framework to implement 
\dstreams . Both these systems are implemented in Scala~\cite{scala} (a language based 
on the Java Virtual Machine), which allows them integrate well with existing Java and 
Scala libraries for linear algebra, machine learning, etc. Furthermore, the compact 
syntax of the Scala language hides all the complexities of distribution, replication and 
data access pattern behind an intuitive programming interface. 
A relevant portion of the code of the algorithm is provided in Appendix B. This code instantiates a D-Stream with the raw data and derives some other D-Streams 
that correspond to each step of the algorithm. As can be seen, this code leverages
the functional API of Spark and Scala to express stream transformations in a very natural way. 
\sys can scale to 
hundreds of cores while achieving latencies as low as hundreds of milliseconds. We 
use this system to implement our traffic estimation algorithms, which we shall 
explain next.

%% file: model.tex
\section{Scalable traffic estimation from streaming data}

\label{sec:model}

We now explain the relevance of D-Streams with a use case of large-scale,
low-latency state estimation: vehicular traffic estimation. The goal of
the traffic estimation algorithm is to infer the travel times of each
link of an arterial road network, given periodic GPS readings from
vehicles moving through the network. We will describe in this section
our estimation framework. We will discuss the performance gains obtained
by using D-Streams in Section \ref{sec:experiments}.

We define the road network as a graph $\mathcal{D}=(\mathcal{V},\mathcal{E})$,
where the set $\mathcal{E}$ will be referred to as the ``links''
of the road network (streets) and $\mathcal{L}$ as the ``nodes''
(road intersections). For each link $l\in\mathcal{E}$, the algorithm
outputs $X_{l}^{t}$, the time it takes at time index $t$ to traverse
link $l$. This time is described as a probability distribution parametrized
by a vector $\nu_{l}$. Our goal is then to estimate $X^{t}$, the
joint distribution of all link travel times across all links in $\mathcal{E}$,
for each time index $t$. We assume that the traffic is varying slowly
enough that it can be considered a steady state between each evaluation:
our algorithm will consider that all the observations between two
consecutive time steps have been generated according to the same state. To simplify
notations, we will consider a single time interval and drop the reference
to time: the joint distribution of travel time is the multidimensional variable $X$.

We will first give an overview of the GPS data that is commercially available today,
and an algorithm that converts raw GPS points to map-matched trajectories
with high accuracy: the \emph{Path Inference Filter} (PIF) 
\cite{hunter12pathinference}. We will then present our modeling approach to infer the traffic conditions
from these GPS observations. Then we will explain how the \emph{Mobile
Millennium} \cite{mm-socc} pipeline implements this algorithm using a computing cloud
and Streaming Spark as a computation backend.

\subsection{Map-matching GPS probe data with the Path Inference Filter\label{sub:map-matching}}

\begin{figure}
\hfill{}\includegraphics[width=1\columnwidth]{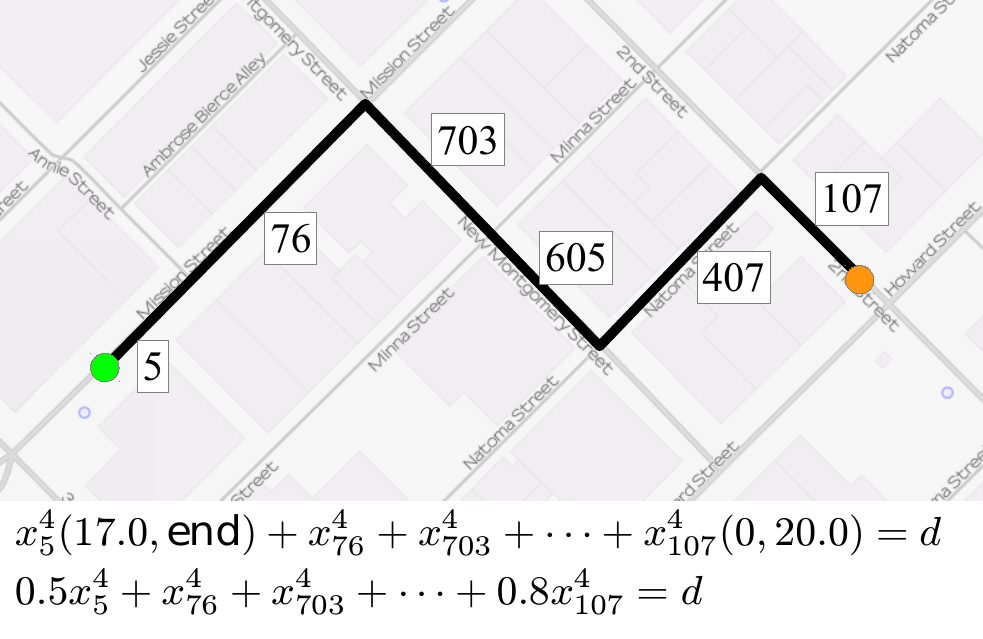}\hfill{}

\caption{Example of observation. The green mark represents an initial GPS reading,
the orange mark represents a subsequent reading. The black line marks the path
of the vehicle, as reconstructed by the Path Inference Filter between the two
GPS points and the numbers are the indexes of each road link covered by
this observation. Given a realization $x^4$ of the travel time distribution at time $t=4$,
all the information on travel times encoded by this observation is summarized in the equation above.
\label{fig:Example-of-observation}}

\end{figure}

In order to reduce power consumption and transmission costs, probe
vehicles do not continuously report their location to the base station.
A high temporal resolution gives access to the complete and precise
trajectory of the vehicle, but this causes the device to consume more
power and communication bandwidth. Also, such data is not available 
at large scale today, except in a very fragment portion of the the private sector.
A low temporal resolution carries
some uncertainty as to which trajectory was followed. In the case
of a high temporal resolution (typically, a frequency greater than
an observation per second), some highly successful methods have been
developed for continuous estimation \cite{thrun2002probabilistic,miwa2004route,du2004lane}.
However, most data collected at large scale today is generated by
commercial fleet vehicles. It is primarily used for tracking the vehicles
and usually has a low temporal resolution (1 to 2 minutes) \cite{navteq,inrix,telenav,cabspotting}.
In the span of a minute, a vehicle in a city can cover several blocks
(see Figure \ref{fig:Example-of-observation} for an example). Information
on the precise path followed by the vehicle is lost. Furthermore,
due to GPS localization errors, recovering the location of a vehicle
that just sent an observation is a non trivial task: there are usually
several streets that could be compatible with any given GPS observation.
Simple deterministic algorithms to reconstruct trajectories fail due
to misprojection or shortcuts. The Path Inference Filter \cite{hunter12pathinference}
is a probabilistic framework that recovers trajectories and road positions
from low-frequency probe data in real time, and in a computationally
efficient manner.

\begin{figure}[tb]
\hfill{}\label{fig:edge-a}(1)\includegraphics[angle=-90,width=1.2in]{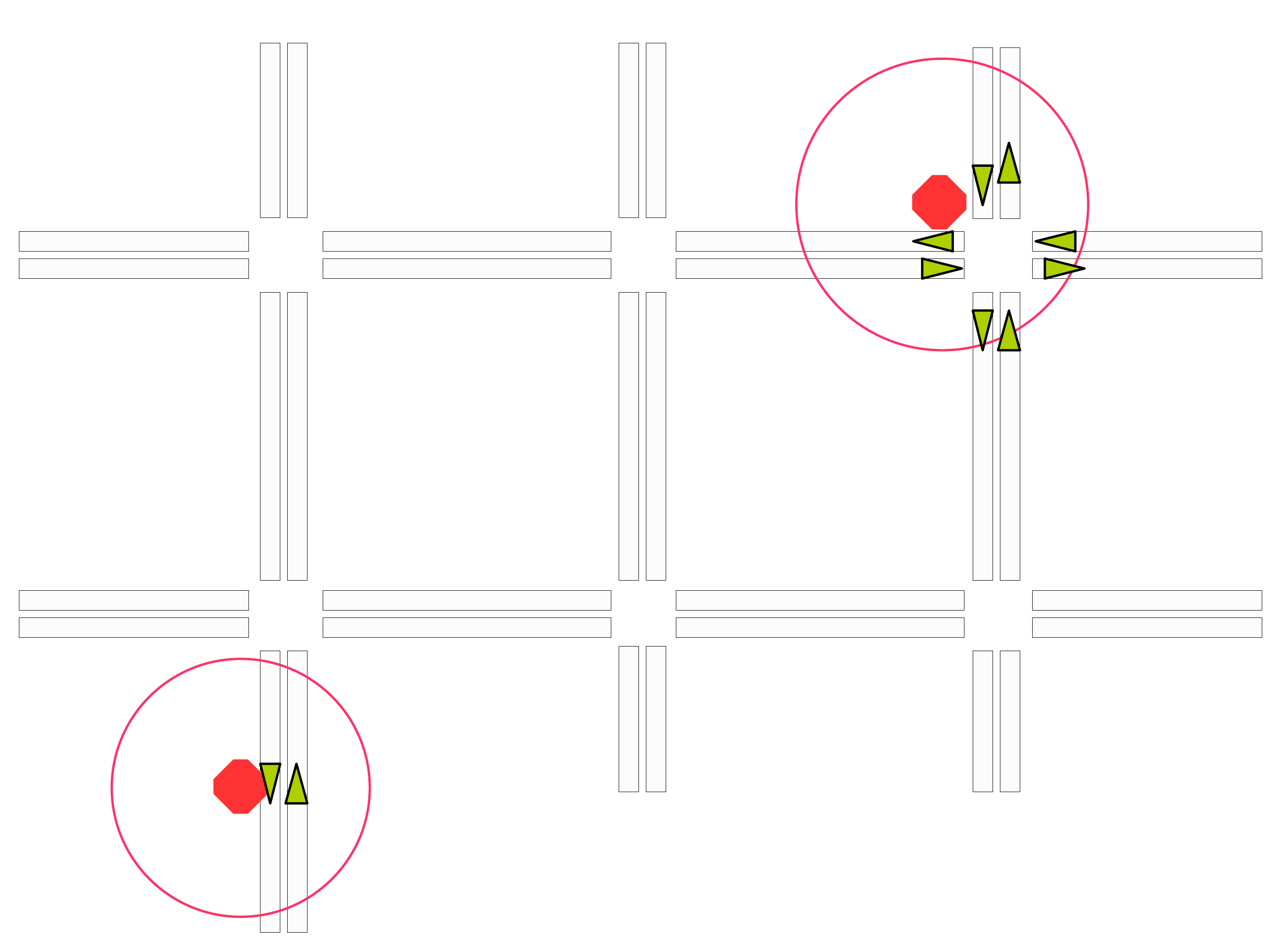}\label{fig:edge-b}\hfill{}(2)\includegraphics[angle=-90,width=1.2in]{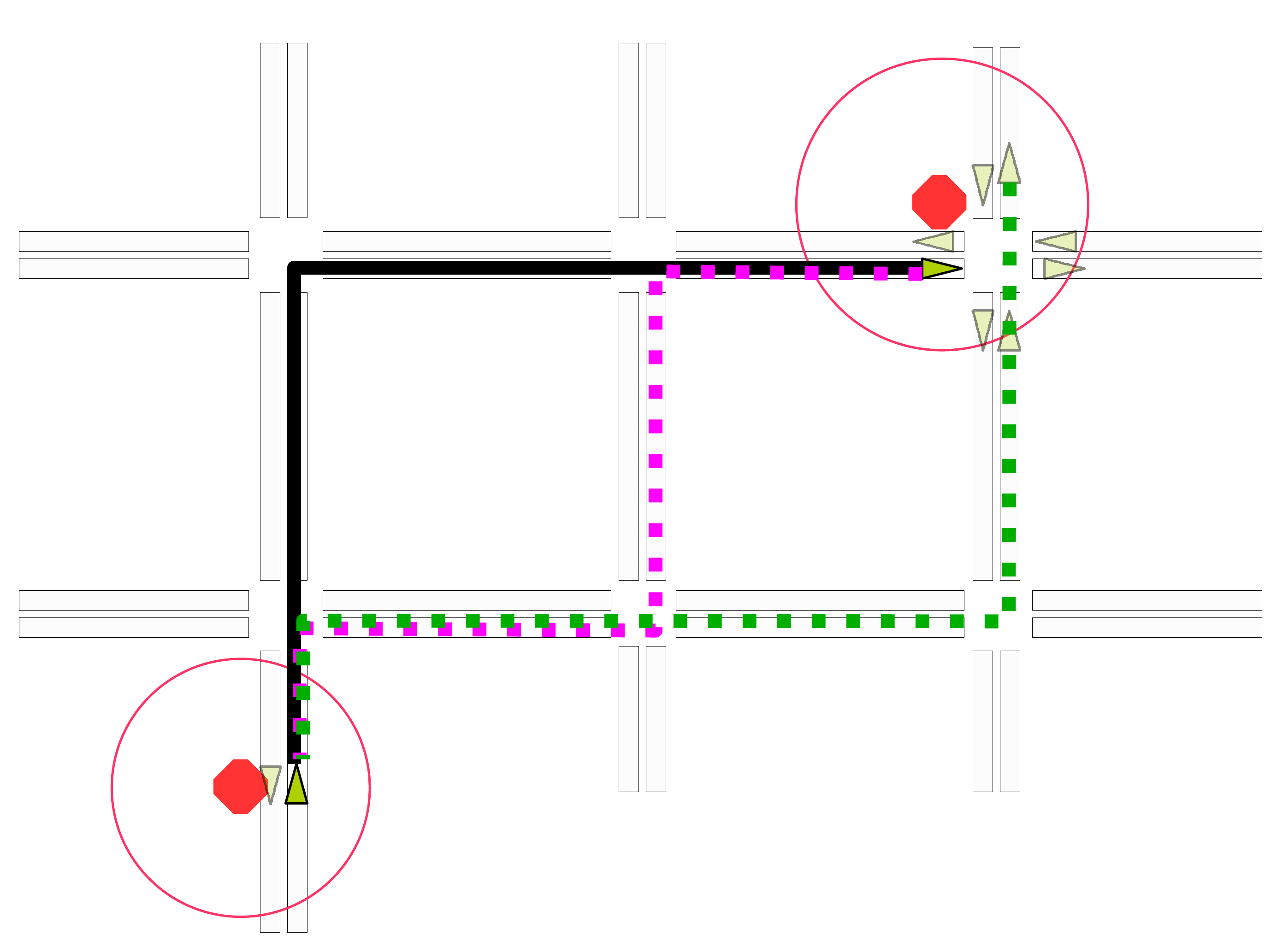}\hfill{}

\hfill{}\label{fig:edge-c}(3)\includegraphics[angle=-90,width=1.2in]{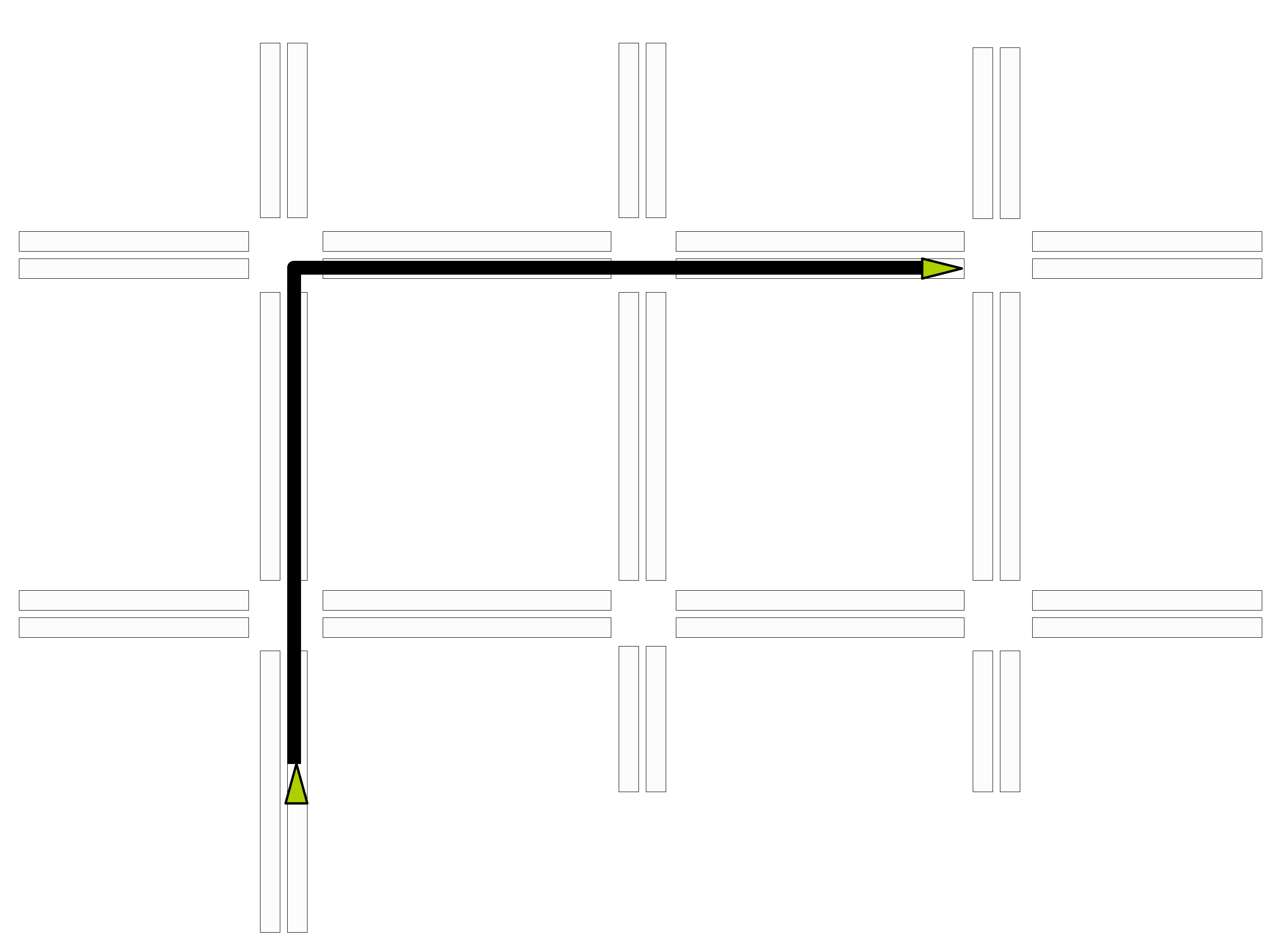}\label{fig:edge-d}\hfill{}(4)\includegraphics[angle=-90,width=1.2in]{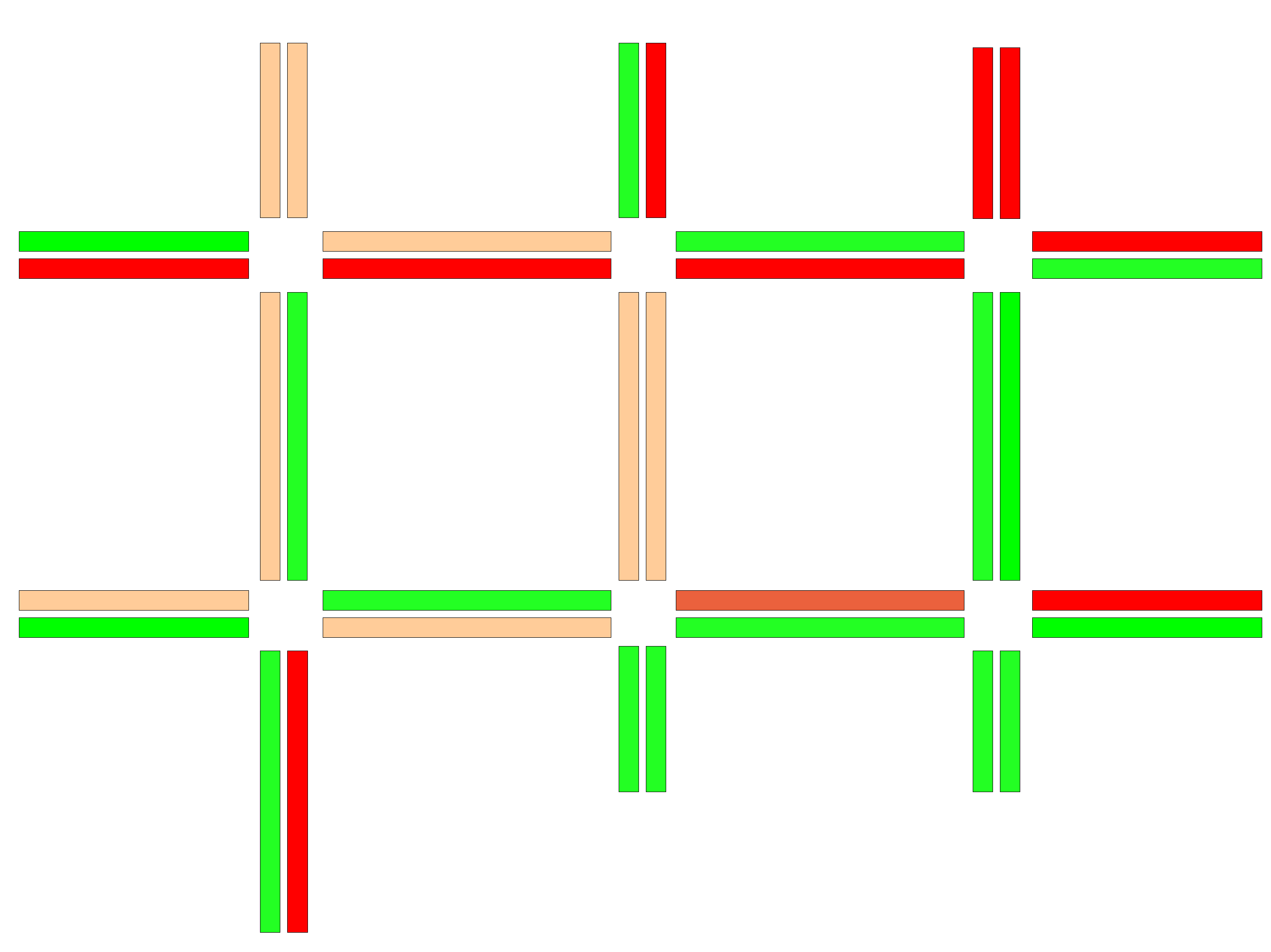}\hfill{}
\caption{Map-matching algorithm: the raw GPS readings a first projected onto
candidate points on the road network (Step 1). Then all feasible paths between each
pair of candidate points are computed (Step 2). A dynamic programing algorithm
then finds the most likely trajectory, using a Conditional Random
Field (Step 3). Trajectory measurements are the input to the Expectation Maximization
algorithm. This algorithm outputs distributions of travel times (Step 4).\label{fig:Map-matching-algorithm}}
\end{figure}

This algorithm first projects the raw points onto candidate projections
on the road network and then builds candidate trajectories to link
these candidate projections. An observation model and a driver model
are then combined in a Conditional Random Field to find the most probable
trajectories, using the Viterbi algorithm. More precisely, the algorithm
performs the following steps:
\begin{itemize}
\item We map each point of raw (and possibly noisy) GPS data to a collection
of nearby \emph{candidate projections} on the road network (Figure
2-1). 
\item For each vehicle, we reconstruct the most likely trajectory using
a Conditional Random Field ~\cite{hunter12pathinference} (Figure
2-2). 
\item Each segment of the trajectory between two GPS points is referred
as an \emph{trajectory measurement} (Figure 2-3). A
trajectory measurement consists in a start time, an end time and a
route on the road network. This route may span multiple road links,
and starts and ends at some offset within some links.
\end{itemize}
At the output of the PIF, we have transformed sequences of GPS readings
into sequences of trajectory readings. These readings are the input
for our travel time estimation algorithm.

\subsection{Fundamental generative model\label{sub:basic-model}}

An example of a trajectory reading is given in Figure \ref{fig:Example-of-observation}.
Estimating the travel time distributions is made difficult by the
fact that we do not observe travel times for individual links. Instead,
each reading only specifies the total travel time for an entire list
of links traveled. We formally describe our estimation task as a maximum
likelihood estimation problem.

We consider one reading, described by an offset on a first road link
$o_{\text{start}}$, an offset on a last link $o_{\text{end}}$, a
list of $m$ visited links $l_{1}\cdots l_{m}$, a start time, and
a travel duration $d$. We simplify the problem by assuming that the
partial travel time from the start of a road link to some offset $o$
is proportional to the offset: the travel time between the start of
the link $l$ and offset $o$ is a probability distribution $X_{\text{partial}}\left(0,o\right)=\frac{o}{L\left(l\right)}X_{l}$
where $L\left(l\right)$ is the length of link $l$. Using this assumption,
we can convert the description of an observation into a vector form:
consider the vector $\alpha\in\mathbb{R}^{n}$ where:
\[
\alpha\left(l_{1}\right)=1-\frac{o_{\text{start}}}{L\left(l_{1}\right)}
\]
\[
\alpha\left(l_{m}\right)=\frac{o_{\text{end}}}{L\left(l_{m}\right)}
\]
\[
\alpha\left(l_{i}\right)=1\text{ for \ensuremath{i\in\left[2...m-1\right]}}
\]
and $\alpha\left(l\right)=0$ for all other links $l$. Thanks to
the proportionality assumption, the observed travel duration $d$
along the path $p$ is a linear combination of linear travel times:
\begin{eqnarray*}
d & = & \sum_{l\in\mathcal{E}}\alpha\left(l\right)x_{l}\\
 & = & \sum_{l\in p}\alpha\left(l\right)x_{l}
\end{eqnarray*}

The vector $\alpha$ is called the \emph{path activation vector} for
this reading. Note that fewer than 10 links are covered in a typical
trajectory measurement, so the path activation vectors are extremely
sparse. We will use this fact to achieve very good scaling of our
algorithm.

For a given time interval, we can completely
represent a trajectory reading by an \emph{observation} $Y=\left(\alpha,d\right)\in\left(\mathbb{R}^{+}\right)^{n}\times\mathbb{R}_{+}^{*}$.
Each observation $Y^{\left(r\right)}=\left(\alpha^{\left(r\right)},D^{\left(r\right)}\right)$
describes the $r$th trajectory's travel time $D^{\left(r\right)}$
and path $\alpha^{\left(r\right)}$ as inferred by earlier stages
of the \mm pipeline. The travel time $D^{\left(r\right)}$ is the
time interval between consecutive GPS observations and is roughly
one minute for our source of data. 

\begin{figure}
\hfill{}\includegraphics[width=0.6\columnwidth]{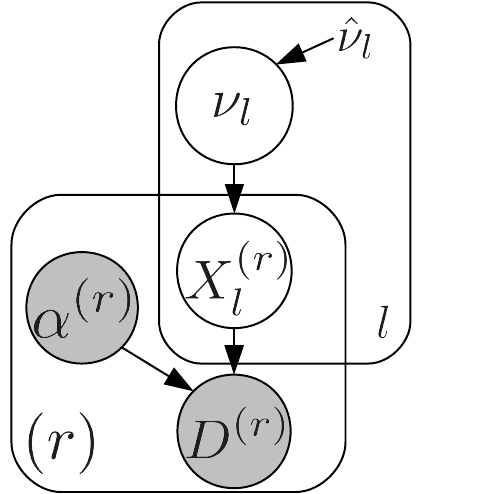}\hfill{}

\caption{\label{fig:Graphical-model}Directed (Bayesian) graph of the travel time model.
Grey nodes are observed variables, white nodes are hidden variables.
The arrows represent conditional dependencies between the variables.
Boxes encode \emph{plates}, i.e. a factorization of repeating variables.
(Right) an expansion of a few elements of the plates.
}
\end{figure}

To make the inference problem tractable, we model the link travel
times for each link $l$ as a Gamma distribution with parameter vector
$\nu_{l}=\left(k_{l},\theta_{l}\right)$, and we assume these distributions
are pairwise independent%
\footnote{We experimented with a few standard distributions from the literature
(Gamma, Normal and Log-normal). Based on our experiments, the Gamma
distribution fit the data best. Computing the most likely Gamma distribution
from a set of samples is more expensive than in the case of the Normal,
but was deemed worthwhile for the added accuracy.%
}. The independence assumption is standard in the transportation literature
\cite{hofleitner_itsc_2011,hofleitnerTRBstatTraffic_2012} and it
also leads to a highly scalable estimation algorithm. We will discuss
the validity of this assumption in Section \ref{sub:experiment-ml}. 

The dependencies between the observations and the parameter vector
$\nu$ can be represented as a Bayesian graphical model, which encodes
all the dependencies between the variables in a very compact form
(Figure \ref{sub:basic-model}). We now formalize the problem of estimating
the set of parameters $\nu=\left(\nu_{l}\right)_{l}\in\mathbb{R}^{n\times2}$
for a set of observations $\left(Y^{\left(r\right)}\right)_{r=1\cdots R}$
as a learning problem. We consider that the the current estimate of
the traffic is completely described by independent Gamma distributions
of the travel times over each road link. These travel times are (indirectly)
observed through a set of observations $Y^{\left(r\right)}=\left(\alpha^{\left(r\right)},d^{\left(r\right)}\right)$.
The set of parameters that maximizes the likelihood of these observations
is solution to the \emph{maximum likelihood problem}:
\begin{eqnarray}
\max_{\nu}\,\, ll\left(Y;\nu\right) & = & \sum_{r}\log\pi\left(D^{\left(r\right)}\bigg|\alpha^{\left(r\right)};\nu\right)\label{eq:max-ll}
\end{eqnarray}
with $\pi\left(D^{\left(r\right)}\bigg|\alpha^{\left(r\right)};\nu\right)$
the probability of observing the duration $d=\sum_{l}\left(\alpha\left(l\right)\right)x_{l}$
when $x_{l}$ is generated according to the distribution $\pi\left(\cdot;\nu_{l}\right)$.
This likelihood can be decomposed using the relations of independence
between variables:
\begin{flalign*}
\pi\left(D^{\left(r\right)}\bigg|\alpha^{\left(r\right)};\nu\right)=\int_{X}\pi\left(D^{\left(r\right)}\bigg|X,\alpha^{\left(i\right)}\right)\pi\left(X;\nu\right)\text{d}X\\
=\int_{X}\pi\left(D^{\left(r\right)}\bigg|X,\alpha^{\left(i\right)}\right)\left(\prod_{l\,:\,\alpha^{\left(r\right)}\left(l\right)>0}\pi\left(X_{l};\nu_{l}\right)\text{d}X_{l}\right)\\
=\int_{X}\pi\left(D^{\left(r\right)}\bigg|X,\alpha^{\left(i\right)}\right)\left(\prod_{l\,:\,\alpha^{\left(r\right)}\left(l\right)>0}\pi\left(X_{l};\nu_{l}\right)\text{d}X_{l}\right)
\end{flalign*}
Estimating the travel time distributions is made difficult by the
fact that we do not observe travel times $X$ for individual links.
Instead, each observation only specifies the total travel time $D$
for an entire list $\alpha$ of links traveled. To get around this
problem, we use the \emph{Expectation Maximization} (EM) algorithm~\cite{dempster1977maximum,nh99:avoteatjisaov}.
The EM algorithm operates in two phases: In the E-step it considers
each travel time measurement and computes a distribution over
allocations of travel time to each of the links.  In the M-step it
computes the link parameters that maximizes the likelihood of the
travel times for the allocations made in the E-step.  By iterating
this process the EM algorithm converges to a set of link parameters
that are a local maximum of the likelihood of the data. In our
setting, we run the EM algorithm
in an online fashion: for each time step, we use the previous time
step as a value, perform a few (iterations and we monitor the convergence
through the expected complete log-likelihood. This form of online
EM gives good results for our application (Section \ref{sec:experiments}). 

While this model is relatively simple, the use of Gamma distributions
substantially complicates the E step, because sums of independent
Gamma distributions have no simple form. A correct computation of
the expected complete log-likelihood requires the normalization constant
of each distribution $X^{\left(r\right)}|\left(\alpha^{\left(r\right)}\right)^{T}X^{\left(r\right)}=D^{\left(r\right)}$.
Computing this normalization factor for each observation is fairly
expensive (it uses an infinite series expansion) and is critical for
good accuracy in practice. These practical considerations are described
in the next section, and the computational burden justifies the need
for cloud computing.

\subsection{Learning with Gamma distributions\label{sub:learning-with-gamma}}

In the previous section, an observation was defined by a duration $d>0$ and
a (sparse) vector $\alpha\in\left(\mathbb{R}_{+}^{*}\right)^{n}$, with the constraint:
\begin{equation}
d=\alpha^{T}x\label{eq:obs-constraint}
\end{equation}
where $x$ is a vector of unobserved samples from independent Gamma
distributions. The Expectation-Maximization algorithm relies on two
operations: computing the conditional distribution $\pi\left(x|d,\alpha;\nu\right)$
of the vector $x$ when $\alpha$ and $d$ are known and the parameter vector $\nu$ is fixed, and sampling
this distribution, i.e. computing $x$ so that $x_{l}\sim\Gamma\left(k_{l},\theta_{l}\right)$
under the constraint \ref{eq:obs-constraint}. In the case of Gamma
distributions, some formulas exist for these operations. For the sake
of clarity, we only present the main results here, without proof or
background. The reader is invited to refer to Appendix A for a longer
exposure.

\textbf{Sampling from conditional gamma distributions.} Consider a
set of $n$ independent Gamma distributions $T_{i}\sim\Gamma\left({k}_{i},\theta_{i}\right)$ with $k\in\left(\mathbb{R}_{+}^{*}\right)^{n}$ and $\theta\in\left(\mathbb{R}_{+}^{*}\right)^{n}$,
a $n$-dimensional vector of positive numbers $\alpha\in\left(\mathbb{R}_{+}^{*}\right)^{n}$
and $d>0$%
\footnote{The function $\Gamma\left(\cdot,\cdot\right)$ will refer to the Gamma
distribution and the function $\Gamma\left(\cdot\right)$ will refer
to the Gamma function.%
}. Call $T$ the joint distribution of all $T_i$s. The purpose of this paragraph is to present some practical formulas
to sample and compute the density function of the conditional distribution:

\[
Z\sim T\bigg|\sum_{i}\alpha_{i}T_{i}=d
\]

We define this distribution over the $n$-dimensional simplex $\mathcal{S}_{\alpha,d}$:
\[
\mathcal{S}_{\alpha,d}=\left\{ x\in\left(\mathbb{R}^{+}\right)\bigg|\alpha^{T}x=d\right\}
\]
Using an appropriate probability measure over the set $\mathcal{S}_{\alpha,d}$,
the probability density function of $Z$ is:

\begin{equation}
f\left(z\right)=\frac{1}{n\kappa\left(k,\hat{\theta}\right)}d^{n-1}\frac{\sqrt{\sum_{i}\alpha_{i}^{2}}}{\prod_{i=1}^{n}\alpha_{i}}\prod_{i}f_{\Gamma}\left(\frac{\alpha_{i}z_{i}}{d};k_{i},\hat{\theta}_{i}\right)
\label{eq:simplex-pdf}
\end{equation}
with:
\[
\hat{\theta}_{i}=d^{-1}\alpha_{i}\theta_{i}
\]
and $f_{\Gamma}$ the density function of the Gamma distribution:
$f_{\Gamma}\left(x;\tilde{k},\tilde{\theta}\right)=\Gamma\left(\tilde{k}\right)^{-1}\theta^{-\tilde{k}}x^{\tilde{k}-1}e^{-\tilde{\theta}^{-1}x}$.
 The normalization factor $\kappa\left({k},\hat{\theta}\right)$
can be described by an infinite series, and is a straightforward adaptation
of a result from Moshopoulos \cite{Moschopoulos}.
The proof is provided in Appendix A and uses a new distribution, called
the \emph{Gamma-Dirichlet distribution}, that generalizes the Dirichlet
distribution.

\begin{algorithm}
Given $\alpha\in\left(\mathbb{R}_{+}^{*}\right)^{n}$
and $d>0$.

Generate $n$ independent samples $a_{i}\sim\Gamma\left({k}_{i},d^{-1}\alpha_{i}\theta_{i}\right)$

$z_{i}=d\alpha_{i}^{-1}\frac{a_{i}}{\sum_{k}a_{k}}$

Then $z\sim Z|\alpha^{T}Z=d$ 

\caption{\label{alg:Sampler-for-Gamma}Sampler for Gamma distributions conditioned
on a hyperplane}
\end{algorithm}

\textbf{Sampling algorithm.} There happens to exist a straightforward
procedure to sample values from the conditional Gamma distribution
$Z$, presented in Algorithm \ref{alg:Sampler-for-Gamma}. A proof
that this algorithm gives the intended result is also given in Appendix
A.

\subsection{Overview of the MM arterial pipeline\label{sub:mm-pipeline}}

The \mm~system incorporates a complete pipeline for receiving probe
data, filtering it, distributing it to estimation engines and displaying
it, all in in real-time.
This software stack, written in Java and Scala, evaluates \emph{probabilistic
distribution of travel times} over the road links, and uses as input
the \emph{sparse, noisy} GPS measurements from probe vehicles.

\begin{figure}[tb]
\includegraphics[width=3.35in]{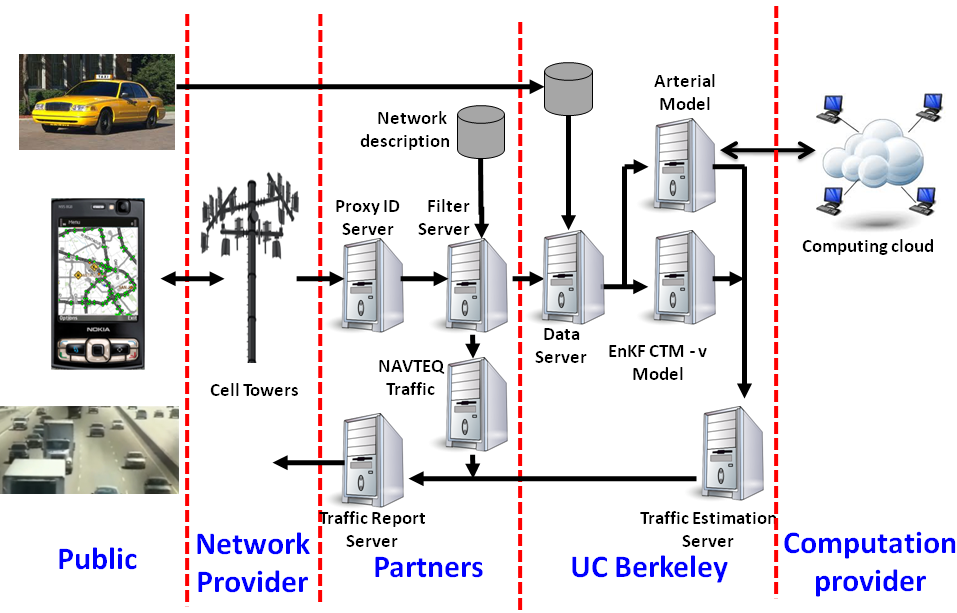} \caption{
Schematic architecture of the \mm system.
\label{fig:system_schema}}
\vspace*{-1em}
\end{figure}

The most computation-intensive parts of this pipeline have all been
ported to a cloud environment. We briefly describe the operations
of the pipeline, pictured in Figure~\ref{fig:system_schema}.

The observations are grouped into time intervals and sent to a traffic
estimation engine, which runs the learning algorithm described in
the next section and returns distributions of travel times for each
link (Figure~\ref{fig:Map-matching-algorithm}). 

The travel time distributions are then stored and broadcast to clients
and to a web interface. Examples of means of travel times are 
shown in Figure~\ref{fig:example-output}. 

It is important to point out that \emph{Mobile Millennium} is intended
to work at the scale of large metropolitan areas. The road network
considered in this work is a real road network (a large portion of
San Francisco downtown and of the greater Bay Area, comprising 506,685
road links) and the data is collected from the field (as opposed to
simulated). A consequence of this setting is the scalability requirement
for the traffic algorithms we employ. Thus, from the outset, our research
has focused on designing algorithms that could work for large urban
areas with hundreds of thousands of links and millions of observations.

Our algorithm needs to run expensive computations in an iterative
fashion on incoming streaming data. As such, it is a perfect candidate
for \dstreams. We now describe how we parallelized the EM algorithm
using Spark's implementation of \dstreams.

\begin{figure}
\hfill{}\includegraphics[width=0.8\columnwidth]{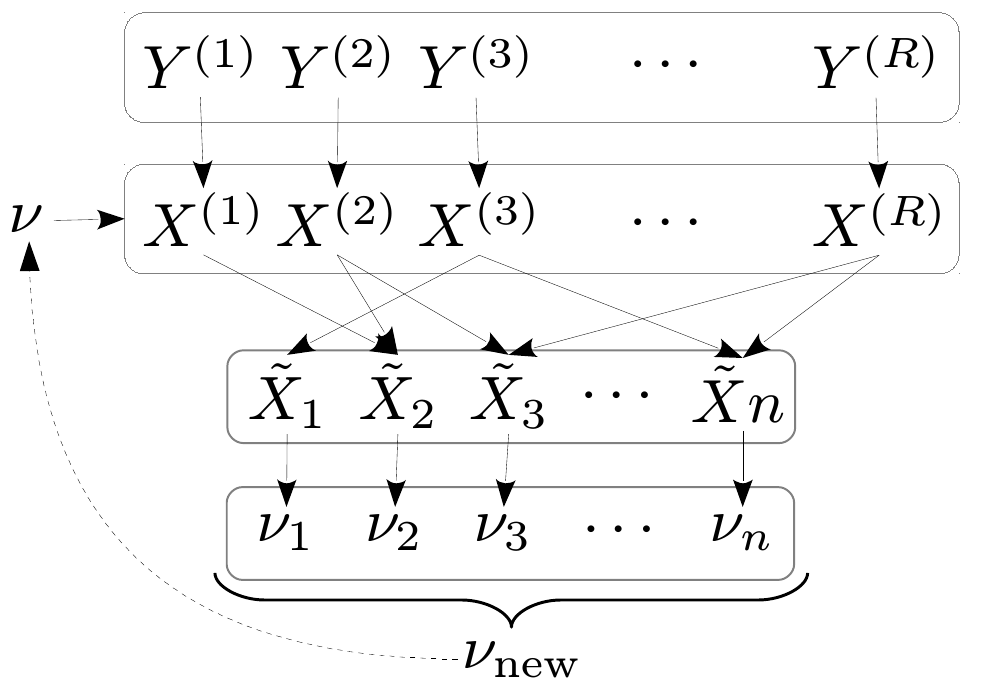}\hfill{}

\caption{System workflow of the EM algorithm. 
In the E-step, we generate per-link travel time samples
from whole observations; specifically, for each observation $Y^{\left(r\right)}=\left(\alpha^{\left(r\right)},d^{\left(r\right)}\right)$,
we produce a set of $U$ weighted samples $\mathbf{X}^{\left(r\right)}=\left\{ \left(x^{\left(r,u\right)},w^{\left(r,u\right)}\right)\right\} _{u=1\cdots U}$,
each sample $x^{\left(r,u\right)}$ produced by randomly dividing
travel time $d^{\left(r\right)}$ among its constituent links (producing
a travel time $x_{l_{i}}^{\left(r,u\right)}$ for each edge $l_{i}\in\alpha^{\left(r\right)}$).
We assign a weight $w^{\left(r,u\right)}$ as the likelihood of travel
time $x^{\left(r,u\right)}$ according to the current distribution
parameters $\nu$. In the shuffle step, we regroup the samples $\mathbf{X}^{\left(r\right)}$
by link, so that each link $l$ now has samples $\tilde{\mathbf{X}}_{l}=\left\{ \left(s_{l}^{\left(r,u\right)},w_{l}^{\left(r,u\right)}\right)\right\} _{r,u}$
from all the observations $r$ that go over it. In the M-step, we
recompute the parameters $\nu_{l}$ to fit link $l$'s travel time
distribution to the samples $\tilde{\mathbf{X}}_{l}$. 
\label{fig:em_flow}}

\end{figure}

Figure~\ref{fig:em_flow} shows the data flow in the algorithm in
more detail. In the E-step, we generate per-link travel time samples
from whole observations; specifically, for each observation $Y^{\left(r\right)}=\left(\alpha^{\left(r\right)},d^{\left(r\right)}\right)$,
we produce a set of $U$ weighted samples $\mathbf{X}^{\left(r\right)}=\left\{ \left(x^{\left(r,u\right)},w^{\left(r,u\right)}\right)\right\} _{u=1\cdots U}$,
each sample $x^{\left(r,u\right)}$ produced by randomly dividing
travel time $d^{\left(r\right)}$ among its constituent links (producing
a travel time $x_{l_{i}}^{\left(r,u\right)}$ for each edge $l_{i}\in\alpha^{\left(r\right)}$).
We assign a weight $w^{\left(r,u\right)}$ as the likelihood of travel
time $x^{\left(r,u\right)}$ according to the current distribution
parameters $\nu$. In the shuffle step, we regroup the samples $\mathbf{X}^{\left(r\right)}$
by link, so that each link $l$ now has samples $\tilde{\mathbf{X}}_{l}=\left\{ \left(s_{l}^{\left(r,u\right)},w_{l}^{\left(r,u\right)}\right)\right\} _{r,u}$
from all the observations $r$ that go over it. In the M-step, we
recompute the parameters $\nu_{l}$ to fit link $l$'s travel time
distribution to the samples $\tilde{\mathbf{X}}_{l}$. 

We note that our EM algorithm is representative of a large class of
iterative machine learning algorithms, including clustering, classification,
and regression methods, for which popular cloud programming frameworks
like Hadoop and Dryad are often inefficient \cite{twister,low2010graphlab}.
Our lessons with Streaming Spark are likely applicable to these applications
too.

%% file: experiments.tex
\section{A case study: taxis in the San Francisco Bay Area}

\label{sec:experiments}

Having now described an algorithm for computing travel time distributions
in real time on a road network, we describe our validation experiments.
These experiments explore two settings:
\begin{itemize}
\item The raw performance of the machine learning algorithm, given a limited
amount of data and a computational budget,
\item The performance of the Streaming Spark framework in distributing computations
across a cluster, and the computational performance improvement gained
by additional hardware.
\end{itemize}
The performance of the algorithm is computed by asking the model to
give travel time distributions on unseen trajectories, slightly in
the future. The observed travel time of the trajectory is then compared
with the distribution provided by the model. We measured the L1 and
L2 losses between the observed travel time and the distribution mean,
and the likelihood of the observed travel time with respect to the
predicted travel time distribution. This is done with different amount
of data and different time horizons.

\begin{figure}[tb]
\hfill{}\includegraphics[width=3.3in]{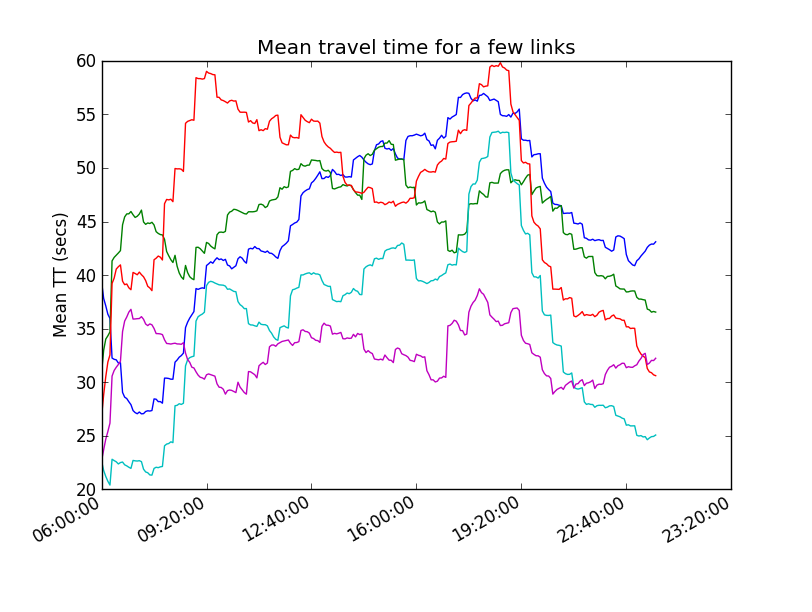}\hfill{}
\caption{An example output of traffic estimates from our algorithm: mean travel
times on different road links in the business district of San Francisco
during a complete day.}

\label{fig:example-output} 
\end{figure}

The computational efficiency of the algorithm is validated in two
steps. First, we demonstrate that our algorithm scales well: given
twice as many computation nodes, it perform the same task about twice
as fast. We also see that this algorithm is bounded by computations.
Then, we demonstrate that it can sustain massive data flow rates under
strict scheduling constraints: we fix a completion time of a few seconds
for each time step, and we find the maximum flow rate under a given
computational budget.

\subsection{Taxis in San Francisco}

\label{sub:experiment-data}

\begin{figure}
\hfill{}\includegraphics[width=0.9\columnwidth]{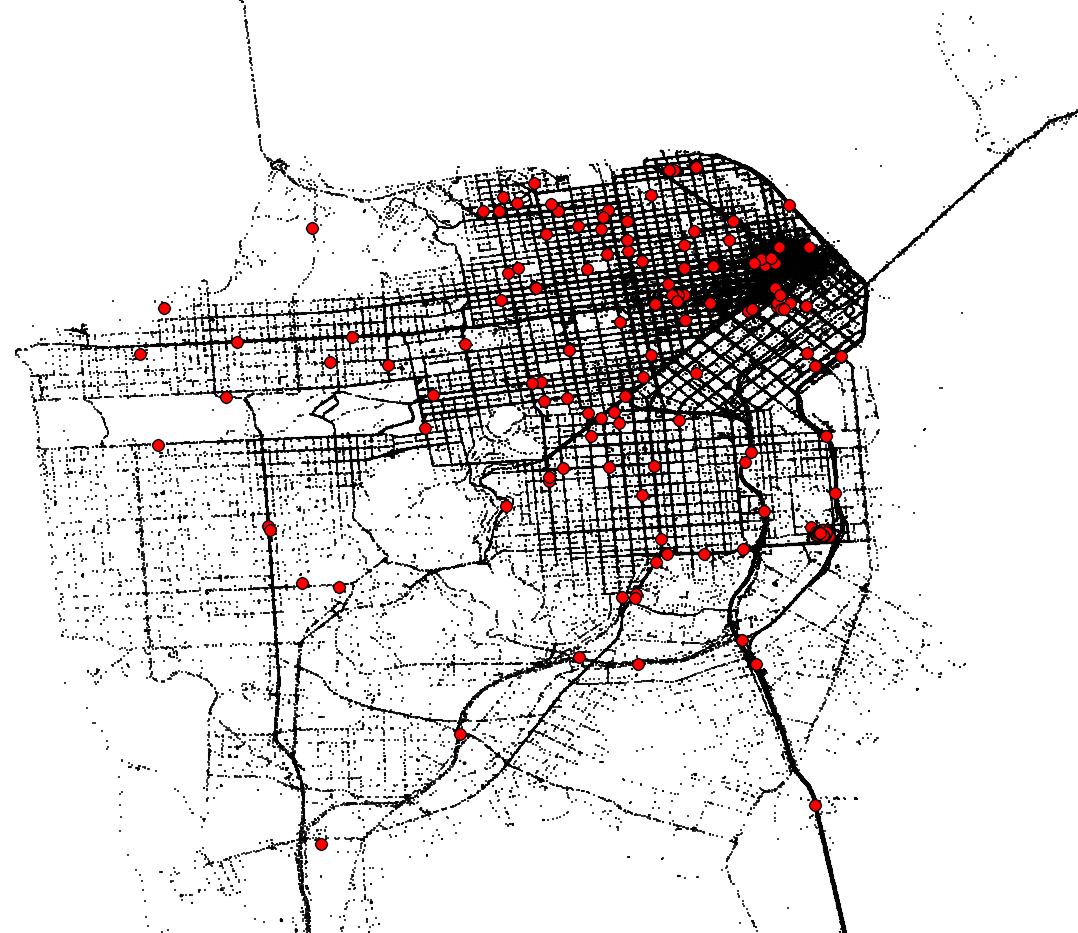}\hfill{}

\caption{An example of dataset available to \emph{Mobile Millennium} and processed
by the path inference filter: taxicabs in San Francisco from the Cabspotting
program. Large circles in red show the position of the taxis at a
given time and small dots (in black) show past positions (during the
last five hours) of the fleet. The position of each vehicle is observed
every minute.\label{fig:cabspotting-example}}
\end{figure}

Our implementation was run on a road network that corresponds to the
greater San Francisco Bay Area (506,685 road links), using some taxi
data provided by the Cabspotting project \cite{cabspotting}. This
dataset contains GPS samples of a few thousand taxicabs emitted every
minute, for more than a year. All in all, it represents hundred of
millions of GPS points. We ran our algorithm on a typical day (August
12th 2010, a Tuesday) with different settings. An example of input
data is given in Figure~\ref{fig:cabspotting-example}. A typical
output of travel times provided by the algorithm is given in Figure
\ref{fig:example-output}.

\subsection{Good scalability results using a large cluster}

\label{sub:experiment-scalability}

In this section, we evaluate how much the cloud implementation helped
with scaling the \mm EM traffic estimation algorithm. Distributing
the computation across machines provides a twofold advantage: each
machine can perform computations in parallel, and the overall amount
of memory available is much greater. 

\begin{figure}
\hfill{}\includegraphics[width=0.8\columnwidth]{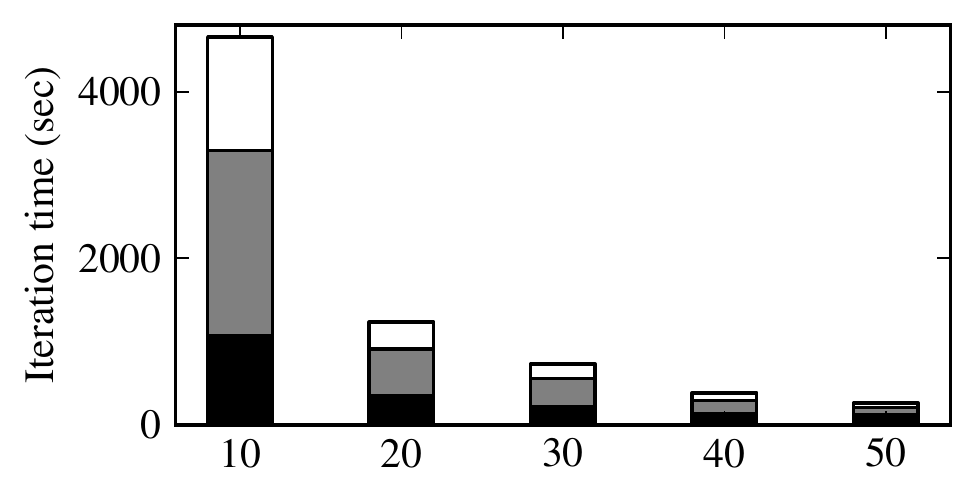}\hfill{}

\hfill{}\includegraphics[width=0.8\columnwidth]{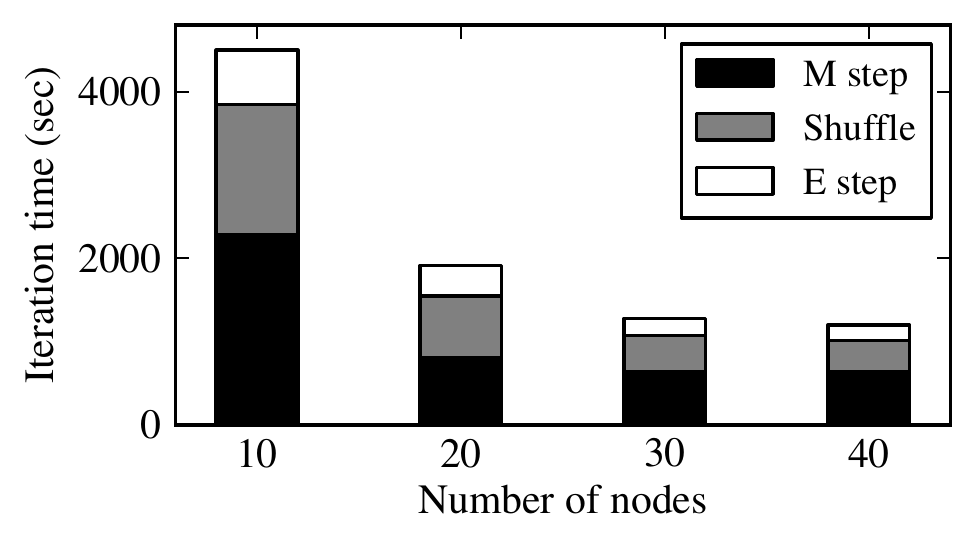}\hfill{}

\caption{Experiments with Spark to build historical estimates of traffic, on
NERSC (top) and Amazon EC2 (bottom).\label{fig:Experiments-with-Spark}\protect \\
}
\end{figure}

\textbf{Scaling.} First, we evaluated how the runtime performance of
the EM job could improve with an increasing number of nodes/cores.
The job was to learn some historical traffic estimate for San Francisco
downtown for a half-hour time-slice, using a large portion of the
data split in one-hour intervals. This data included 259,215 observed
trajectories, and the network consisted of 15,398 road links. We ran
the experiment on two cloud platforms: the first was using Amazon
EC2 \texttt{m1.large} instances with 2 cores per node, and the second
was a cloud managed by the \emph{National Energy Research Scientific Computing
Center} (NERSC) with 4 cores per node. Figure \ref{fig:Experiments-with-Spark}
(bottom) shows near-linear scaling on EC2 until 80--160 cores. Figure
\ref{fig:Experiments-with-Spark} (top) shows near-linear scaling
for all the NERSC experiments. The limiting factor for EC2 seems to
have been network performance. In particular, some tasks were lost
due to repeated connection timeouts.

\begin{figure}
\hfill{}\includegraphics[width=0.7\columnwidth]{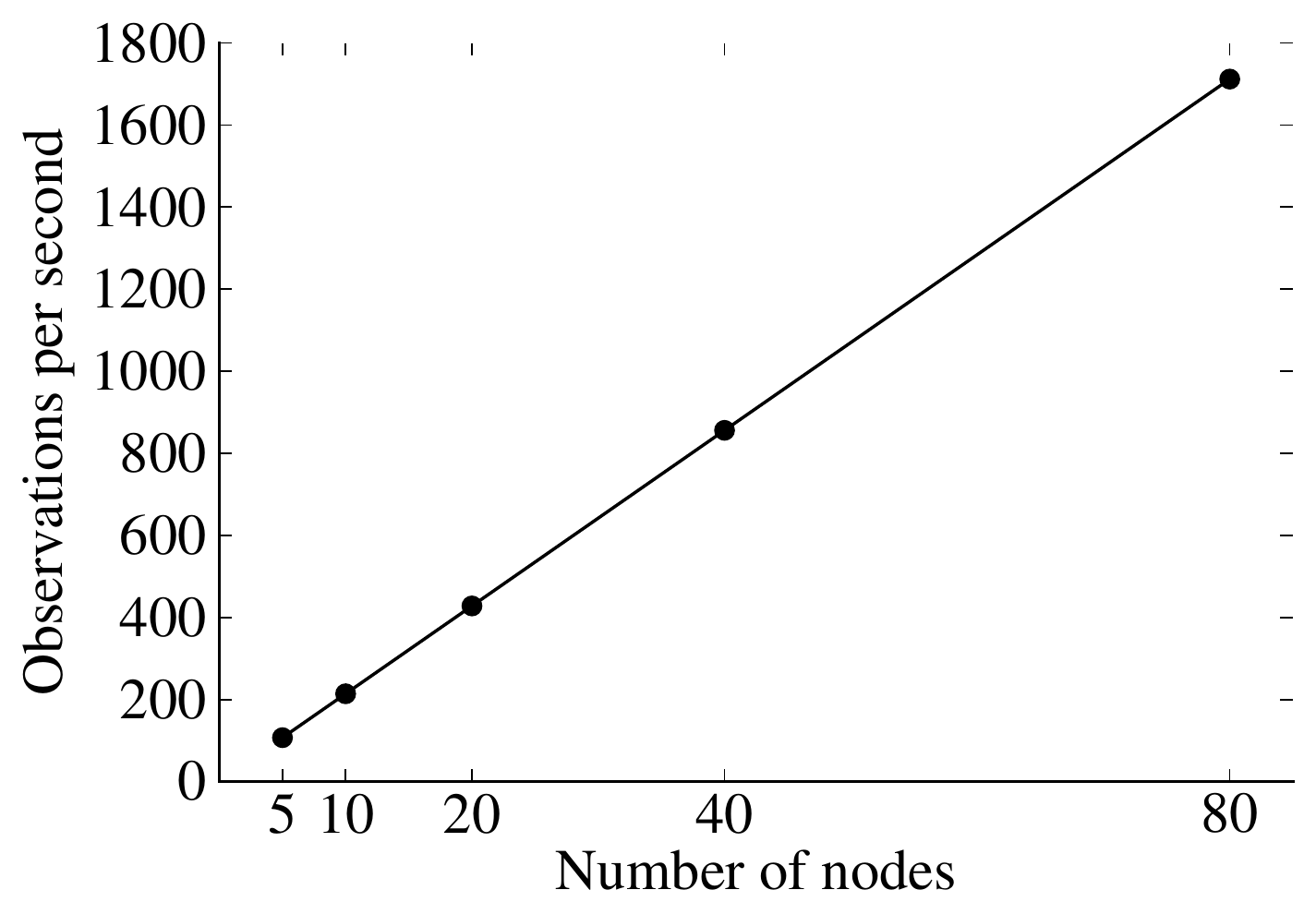}\hfill{}

\caption{Experiments with Streaming Spark: rate of observation processing for
different cluster sizes. \label{fig:Experiments-with-Streaming}}
\end{figure}

\textbf{Scaling on Streaming Spark.} After having found the bottlenecks
in the Spark program, we wrote another version in Streaming Spark.
The two programs are strikingly similar (see program listing in Appendix
B). We then benchmarked the application. We ported this application
to Spark Streaming using an online version of the EM algorithm that
merges in new data every fi{}ve seconds. The implementation was about
200 lines of Spark Streaming code, and wrapped the existing map and
reduce functions in the offl{}ine program. In addition, we found that
only using the real-time data could cause overfi{}tting, because the
data received in fi{}ve seconds is so sparse. We took advantage of
D-Streams to also combine this data with historical data from the
same time during the past 10 days to resolve this problem. Figure
\ref{fig:Experiments-with-Streaming} shows the performance of the
algorithm on up to 80 quad-core EC2 nodes. The algorithm scales almost
perfectly, largely because it is so CPU-bound, and provides answers
an order of magnitude faster than the previous implementation.

\subsection{Our algorithm can be adjusted for trade-offs between amount of data,
computational resources and quality of the output}

\label{sub:experiment-ml}We now study the accuracy of our algorithm
in estimating the traffic. Even if we receive a large number of observations
per day, this number is not sufficient to cover properly in real time
all the road network: indeed, some sections of the road network are
much less traveled than the busy downtown areas. We use several strategies
to mitigate this spatial discrepancy:
\begin{itemize}
\item We use a prior on the Gamma distribution, itself a Gamma distribution
since the Gamma is in the exponential family and conjugate with itself.
The parameters of this prior are to 70\% of the speed limit in mean
and 1 minute or 50\% of the travel time in standard deviation, whichever
is greater.
\item We incorporate some data from the same day before the current time
step, weighted by an exponential decay scheme: the traffic in the
arterial network is assumed to change slowly enough.
\item We also incorporate some data from previous days, corresponding to
the same day of the week (Monday, Tuesday, etc.). Traffic is expected
to follow a weekly pattern during the same month.
\end{itemize}
To summarize, a large number of observations are lumped together and
weighted according to the formula:
\[
w=e^{-\Delta t_{\text{day}}^{-1}\left(t_{\text{obs}}-t_{\text{current}}\right)}e^{-\Delta t_{\text{week}}^{-1}\left(week_{\text{obs}}-week_{\text{current}}\right)}
\]
The half-time decaying factors $\Delta t_{\text{day}}$ and $\Delta t_{\text{week}}$
are set so that the corresponding weight is 0.2 at the end of the
window.

Since the EM learning algorithm is not linear in the observations, we cannot reduce each
observation to some sufficient statistics. As the algorithm moves
forward in time, each observation will appear at different time steps
with a different weight and needs to be reprocessed. This is a significant
limitation from this approach, but it makes for a good testing ground
of Streaming Spark.

Our EM algorithm can be adjusted in several ways:
\begin{itemize}
\item The number of weeks of data to look back (between 1 and 10)
\item The time window to consider before the current observation (between
20 minutes and 2 hours)
\item The number of samples generated during the E-steps (10-100)
\item The number of EM iterations (1-5)
\item The duration of each time step (5 seconds-15 minutes)
\end{itemize}
The observations we process all have a duration of one minute, but
travel times experienced by users are usually much longer (10 minutes
to a few hours). As such, a good metric for assessing the quality
of a model should not be on predicting travel times for one-minute
observations, but on longer distances. Hour-long travels are very
likely to go be spent mostly on highways, which is not the scope of
this study, and taxicabs usually make small trips (10 to 30 minutes).
This is why we focus our attention to travel times between one minute
(the observations) and 30 minutes (typical durations for taxi rides).
As far as we know, this study of different durations is seldom done
in the study of traffic, which limits any attempt to compare the performance
between different algorithms.

The longer trajectories are obtained from the Path Inference Filter.
They are then cuts into different pieces of the same length (1 minute,
5 minutes, 10 minutes, 20 minutes). Each piece of trajectory is considered
as an independent piece of trajectory for the purpose of travel time
prediction.

We ran the algorithm with 4 different settings:
\begin{itemize}
\item \SBO: the most expensive setting (10 weeks of data, 2 hours of data,
100 EM samples, 5 EM iterations, 15 time steps), used as the baseline
for comparison. Travel time estimates are produced every 20 minutes,
\item \SBA: uses less data (40 minutes of data),
\item \SBB: uses less data (10 days),
\item \SBC: uses the same amount of data, but performs only a single EM
iteration every 4 minutes instead of 5 EM iterations every 20 minutes,
\item \SBD: uses the same amount of data, but generates only 10 EM samples
for each observation
\end{itemize}
For all these experiments, the prior was fixed.

\begin{figure}
\hfill{}\includegraphics[width=0.4\columnwidth]{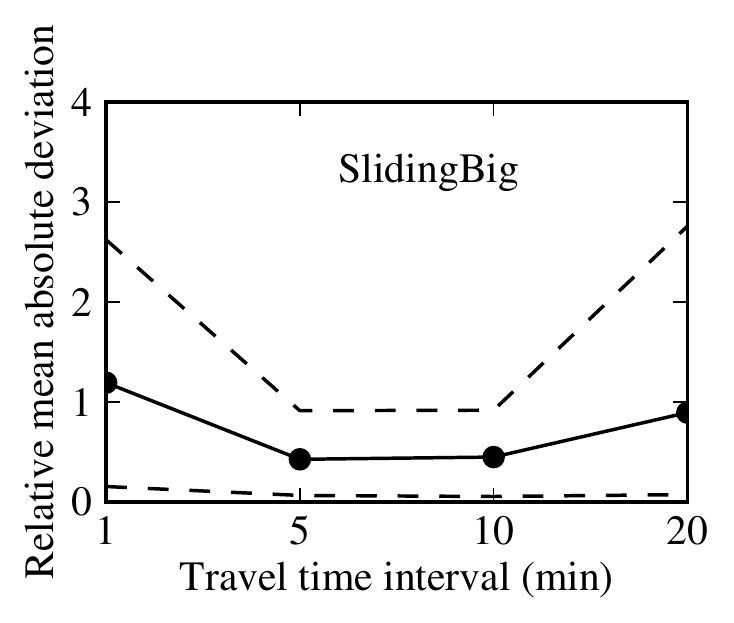}\hfill{}\includegraphics[width=0.4\columnwidth]{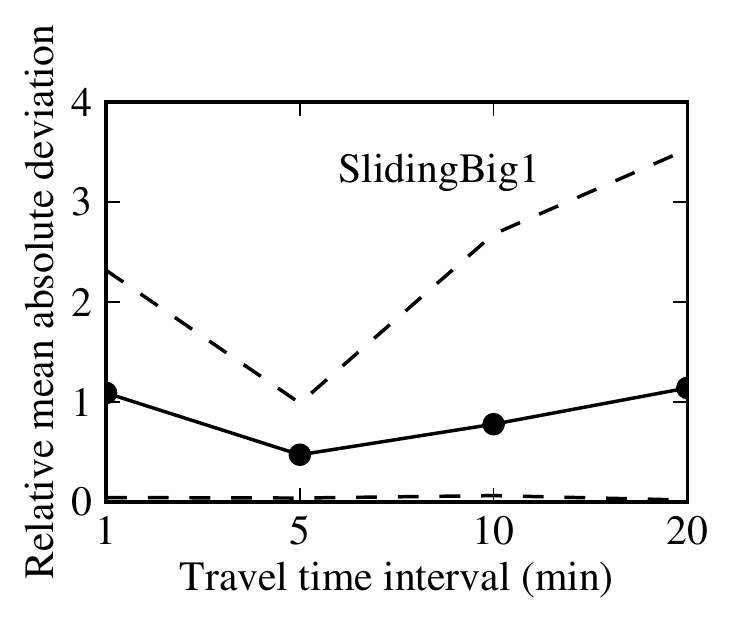}\hfill{}

\hfill{}\includegraphics[width=0.4\columnwidth]{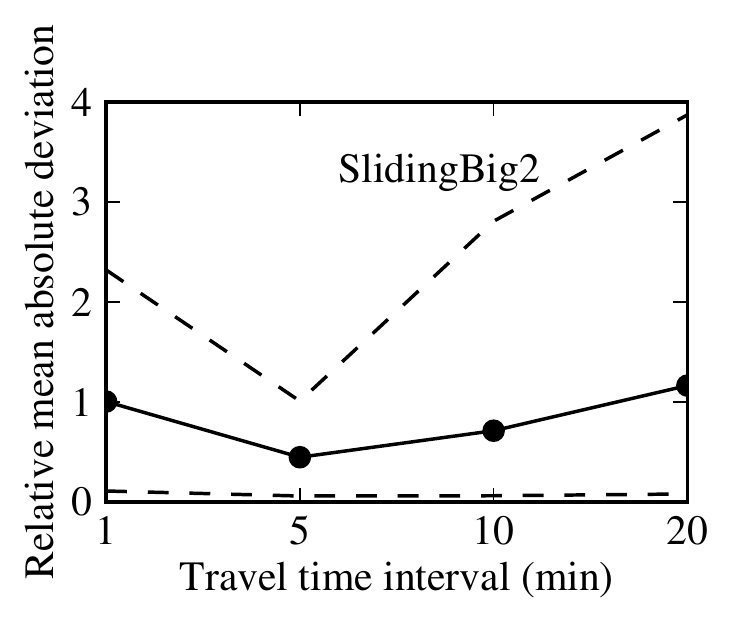}\hfill{}\includegraphics[width=0.4\columnwidth]{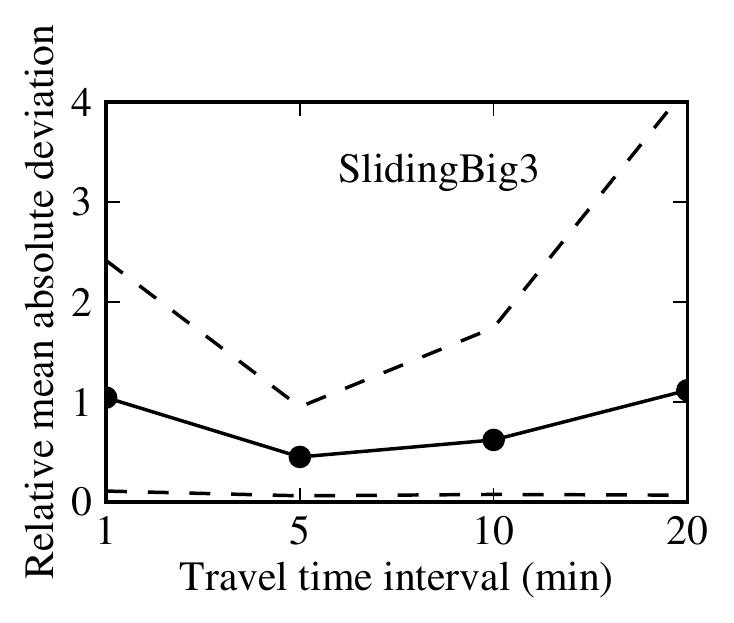}\hfill{}

\hfill{}\includegraphics[width=0.4\columnwidth]{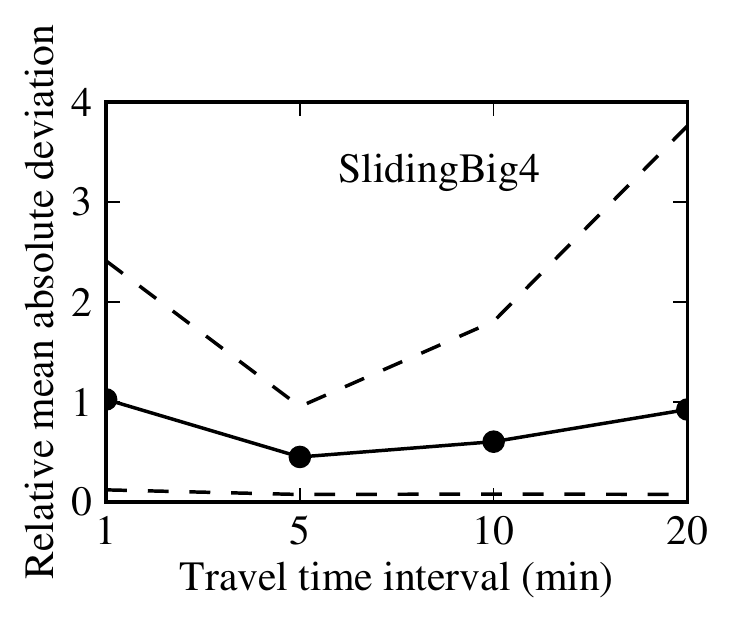}\hfill{}

\caption{L1 residuals for different settings and for different travel times.
The dashed lines indicate the 95\% confidence interval. On the $x$ axis, the travel time of the 
the observations considered for this metric.\label{fig:L1}}
\end{figure}

We now compare the results obtained with the different experiments.
We first turn our attention to the L1 loss in Figure \ref{fig:L1}.
As expected, the best performance is obtained for experiment \SBO,
which uses the most data. Interestingly enough, the best performance
is obtained for travels of medium length (4-11 minutes), and not for
short trajectories. This can be explained by the conversion step that
transforms trajectory readings on partial links into weighted observations
on complete links. The relation between link travel time and location
on a link is more complex than a linear weighting. Nevertheless, the
model gives relatively good performance by this simple transform.
When a vehicle is stopped at a red light, it does not travel along
the link but still has a non-zero travel time. In this case, the weight
of an observation is taken to be half of the total travel time of
the link. In particular, the relative error increases as the duration
(and the length) of travels increases. Performance is not too different
between experiments, which suggests some even smaller amount of data
could be considered.

\begin{figure}
\hfill{}\includegraphics[width=0.4\columnwidth]{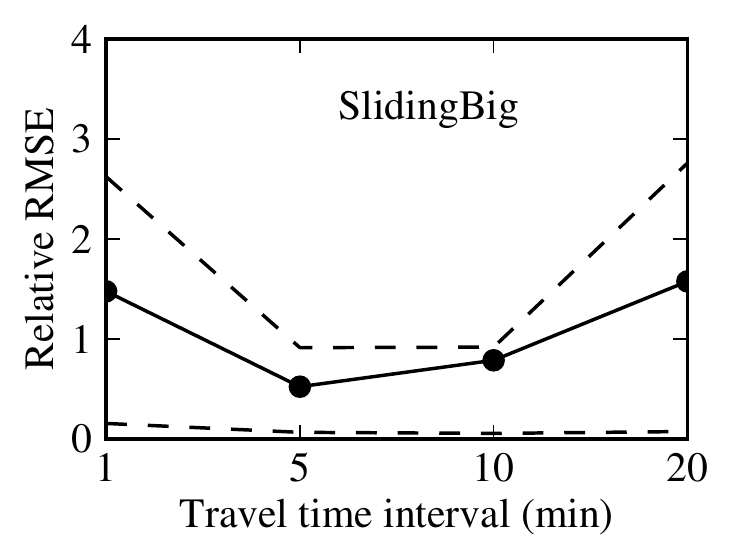}\hfill{}\includegraphics[width=0.4\columnwidth]{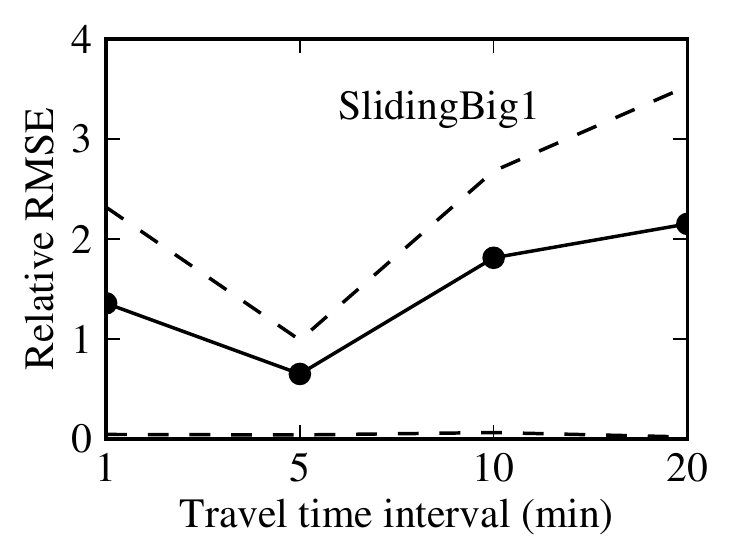}\hfill{}

\hfill{}\includegraphics[width=0.4\columnwidth]{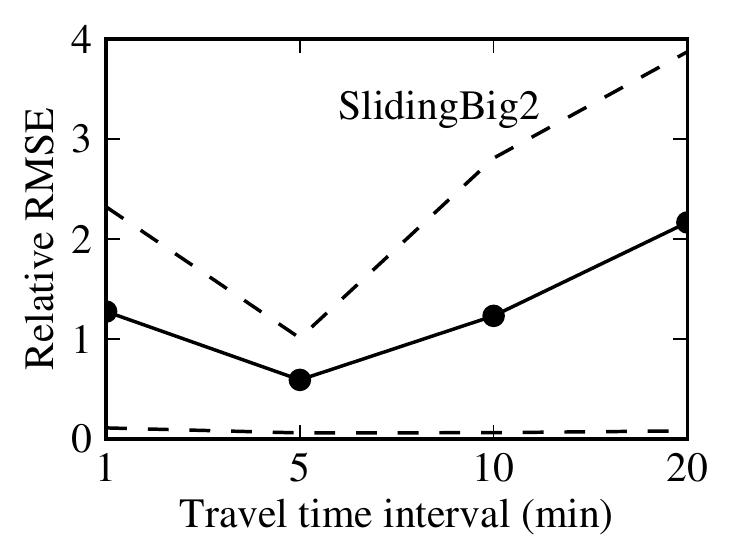}\hfill{}\includegraphics[width=0.4\columnwidth]{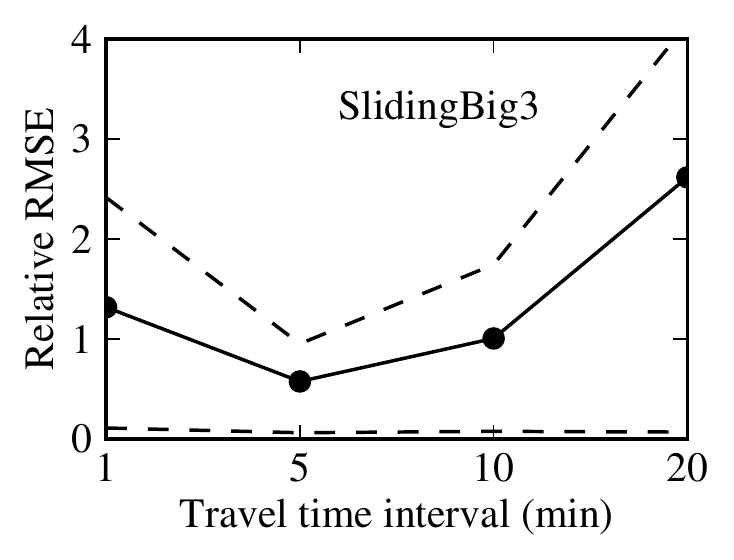}\hfill{}

\hfill{}\includegraphics[width=0.4\columnwidth]{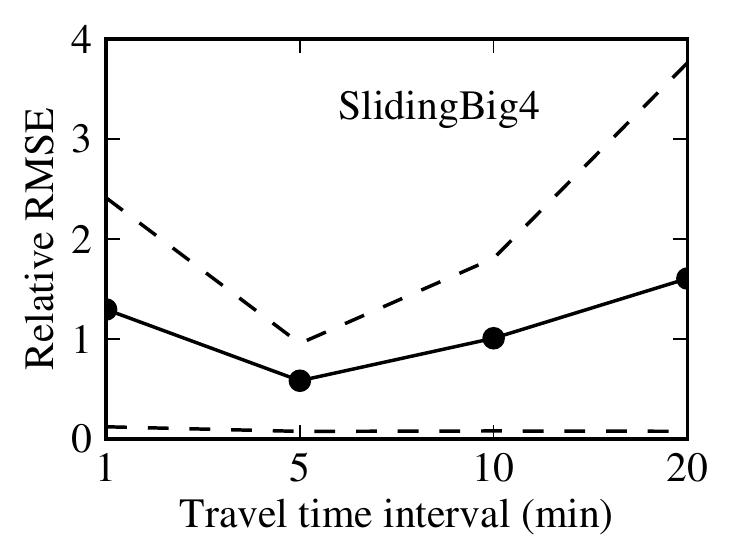}\hfill{}

\caption{L2 residuals for different settings and for different travel times.
The dashed lines indicate the 95\% confidence interval.\label{fig:L2}}
\end{figure}

The results for the L2 loss, presented in Figure \ref{fig:L2}, provide
some similar, if more acute, results. The RMSE is lowest for small
to medium travels (in the range of 3-10 minutes).

\begin{figure}
\hfill{}\includegraphics[width=0.7\columnwidth]{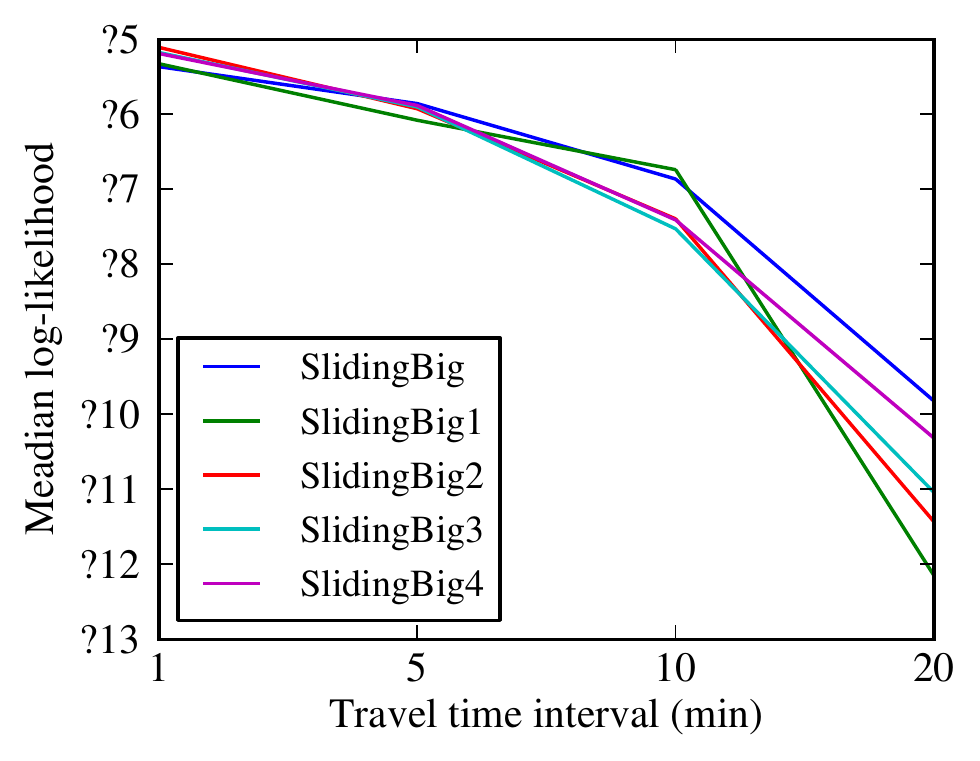}\hfill{}

\caption{Log-likelihood of unobserved trajectories, for different trajectory
lengths and different settings.\label{fig:LL}}
\end{figure}

A probabilistic metric (the log-likelihood) gives a different insight,
as shown in Figure \ref{fig:LL}. The model best explains the data
for very short travel times (similar to what is was trained on) but
its precision falls down as the length of trajectories increases.
All in all, this results should not be unexpected: this model with
independent links cannot take into account the correlations that occur
due to light synchronizations or drivers' behavior. As such, the
probability density of a longer travel rapidly dilutes as the number
of links increases As we saw with our study of L1 and L2 errors, the
mean travel time becomes the only significant value of interest for
longer travel times. In the light of this result, there seems to be
little to gain by modeling travel time with physically realistic,
link-based, independent distributions, as the independence assumption
will strongly weigh on the quality of the travel time for longer travels.
Instead, we recommend focusing effort on simpler models of travel
times that take into account the correlations between links.

%% file: conclusion.tex
\section*{Conclusion}

As datasets grow in size, some new strategies are required to perform
meaningful computations in a short amount of time. We explored using a
new technique, Discretized Streams or D-Streams, that offers some significant
advantages for implementing large-scale state estimation in near-real-time.
D-Streams were implemented in the Spark computing framework. This
approach was validated with a real-life, large-scale estimation problem:
vehicular traffic estimation. Our traffic algorithm is an Expectation-Maximization
algorithm that computes travel time distributions of traffic by incremental
online updates. This algorithm seems to compare favorably with the
state of the art and shows some attractive features from an implementation
perspective. When distributed on a cluster, this algorithm scales
to very large road networks (half a million road links, tens of thousands
of observations per second) and can update traffic state in a few
seconds.

In order to foster research in systems and in traffic, the authors
have released the code of \sys~\cite{spark-code}, the code of the
EM traffic algorithm~\cite{bots-arterial-streaming-code}, and the dataset used for these experiments~\cite{bots-arterial-streaming-code}.

\section*{Acknowledgments}

This research is supported in part by gifts from Google, SAP, Amazon
Web Services, Cloudera, Ericsson, Huawei, IBM, Intel, Mark Logic,
Microsoft, NEC Labs, Network Appliance, Oracle, Splunk and VMWare,
by DARPA (contract \#FA8650-11-C-7136), and by the National Sciences
and Engineering Research Council of Canada. The generous support of
the US Department of Transportation and the California Department
of Transportation is gratefully acknowledged. We also thank Nokia
and NAVTEQ for the ongoing partnership and support through the \emph{Mobile
Millennium} project.

Many thanks to Jack Reilly and Samitha Samarayanake for helpful discussions.
The authors wish to thank Ghada Elabed for her insightful comments on the draft.

%% file: gammadir.tex
\section*{Appendix A: The Gamma-Dirichlet distribution}

We provide proofs and more detailed exposure to the claims made in
Section \ref{sub:basic-model}. We formally introduce a straightforward
generalization of the Dirichlet distribution, which we call the \emph{Gamma-Dirichlet
distribution}%
\footnote{Even if this extension is quite simple, to our knowledge, the following
results have not been presented in the literature so far.%
}. We detail the derivation of the p.d.f. and the sampling procedures
of the conditional distribution $Z|\alpha^{T}Z=t$.

\textbf{The Gamma-Dirichlet distribution.} The regular simplex $\mathcal{S}^{n}\subset\mathbb{R}^{n}$
is the convex hull of the elementary vertices $\left(\mathbf{e}_{i}\right)_{i\in\left[1,n\right]}$.
Given a vector ${k}\in\left(\mathbb{R}^{+}\right)^{n}$ and
a vector $\theta\in\left(\mathbb{R}^{+}\right)^{n}$, we define the
\emph{Gamma-Dirichlet distribution}:
\[
X\sim\Gamma D\left({k},\theta\right)
\]
 with values over the regular simplex $\mathcal{S}^{n}$ as the normalized
sum of $n$ elements drawn from independent Gamma distributions:
\[
X_{i}=\frac{Y_{i}}{\sum_{j}Y_{j}}
\]

with 
\begin{equation}
Y_{i}\sim\Gamma\left({k}_{i},\theta_{i}\right)\label{eq:Y}
\end{equation}
and $Y_{i}$ all pairwise independent. The Gamma-Dirichlet distribution
is a simple generalization of the Dirichlet distribution: if $\theta=a\mathbf{1}$
for some $a>0$, this is the Dirichlet distribution of the $n$th
order. The definition gives a straightforward procedure to sample
some values from $X$. We now present some new results: we give a
formula for the density function that is amenable for computations
and we describe some heuristics that speed up computations by some
significant margin with no significant loss of precision in practice.
Finally, we study the relation between the Gamma-Dirichlet distribution
and the conditional distribution $Z|\alpha^{T}Z=d$, in which a vector
of independent Gamma distributions is jointly constrained on a hyperplane.
Specifically, we show this distribution over $Z$ is equivalent to
the Gamma-Dirichlet distribution.

\textbf{Density.} Note first that one needs to be careful in defining
the underlying $\sigma$-algebra of our probability space, as the
values of $X$ are located in an embedding of $\mathbb{R}^{n}$ of
measure $0$ (a hyperplane). Consider the $n$-dimensional hyperplane
$\mathcal{H}^{n}=\left\{ x|x^{T}\mathbf{1}=1\right\} $. This hyperplane
includes the simplex $\mathcal{S}^{n}$. The Lebesgue measure of this
set in $\mathbb{R}^{n}$ is zero. However, we can consider the Lebesgue
measure $\tilde{\mu}$ defined over $\mathbb{R}^{n-1}$ and the transform:
$\phi:\mathbb{R}^{n-1}\rightarrow\mathcal{H}^{n}$ defined by $\phi\left(u\right)=\left(u^{T}\mathbf{1}-\sum_{j}u_{j}\right)^{T}$.
This transform is a mapping, so it lets us define a new measure $\hat{\mu}$
for the space $\mathcal{H}^{n}$ based on $\tilde{\mu}$. Under this
measure, the measure of the simplex $\mathcal{S}^{n}$ is positive.
Call $\mu$ the measure defined over $\mathcal{S}^{n}$ by $\mu\left(\cdot\right)=\hat{\mu}\left(\mathcal{S}^{n}\right)^{-1}\hat{\mu}\left(\cdot\right)$.
With respect to this measure, the Gamma-Dirichlet distribution has
density function:
\[
f\left(x\right)=\kappa\left({k},\theta\right)^{-1}\prod_{i=1}^{n}f_{\Gamma}\left(x_{i};{k}_{i},\theta_{i}\right)
\]
with $f_{\Gamma}\left(x;k,\theta\right)$ the p.d.f. of the gamma
distribution: $f_{\Gamma}\left(x;k,\theta\right)=\Gamma\left(k\right)^{-1}\theta^{-k}x^{k-1}e^{-\theta^{-1}x}$.
The normalization factor $\kappa\left({k},\theta\right)$ is
defined by an infinite series, based on a result of Moschopoulos \cite{Moschopoulos}:
\[
\kappa\left(k,\theta\right)=\left(\prod_{i=1}^{n}{k}_{i}^{-\theta_{i}}\right)e^{-1/\tilde{k}}\sum_{l=0}^{\infty}\frac{\delta_{l}}{\tilde{k}^{l}\Gamma\left(\tilde{\theta}+l\right)}
\]
with
\[
\tilde{k}=\sum_{i}{k}_{i}
\]
 and $\left(\delta_{j}\right)_{j}$ a series defined by the recursive
formula:
\[
\begin{cases}
\delta_{0}= & 1\\
\delta_{l}= & \frac{1}{l}\sum_{m=0}^{l}\delta_{l}\left(\sum_{i=1}^{n}k_{i}\rho_{i}^{l-m}\right)
\end{cases}
\]

\begin{IEEEproof}
This result can be obtained by first proving the equivalence of the
Gamma-Dirichlet distribution with the conditional distribution $Z|\sum_{j}Z_{j}=1$,
which is done below and does not need the scaling constant $\kappa$.
Now, consider the distribution of the normalization factor $Y=\sum_{j}Y_{j}$.
This variable is a sum of independent Gamma variables, with p.d.f
$f_{Y}$. Then $\kappa\left({k},\theta\right)=f_{Y}\left(1.0\right)$.
An expression for this coefficient in terms of a converging series
of Gamma coefficients is given in \cite{Moschopoulos,Alouini2001}.
\end{IEEEproof}
The computation of $\delta_{i}$ is quadratic in time, which can be
too slow if tens of thousands of coefficients are required before
convergence, as it may happen in our application. We present below
some heuristics to speed up the computations of this sequence.

\textbf{Equivalence with conditional Gammas.} Consider the conditional
distribution $Z_{i}=Y_{i}|\sum_{j}Y_{j}=1$ with $Y_{i}$ defined
in Equation \ref{eq:Y}. The density function of this distribution
is $0$ over $\mathbb{R}^{n}$ nearly everywhere, however it has non-zero
measure over the regular simplex $\mathcal{S}^{n}$. Given a measure
$\mu$ defined over $\mathcal{S}^{n}$, the two distributions are
the same:
\[
\forall x\in\mathcal{S}^{n},\,\, f_{\Gamma D}\left(x\right)=f\left(x\right)
\]

\begin{IEEEproof}
This proof is adapted from a similar proof \cite{devroye1986non}
for the Dirichlet distribution. Using the same notations as above,
define $Y=\sum_{j}Y_{j}$ and $X_{i}=Y_{i}/Y$ for $i\leq n-1$. The
joint density for the $Y$'s is:
\[
f\left(y\right)\propto\prod_{i}y_{i}^{{k}_{i}-1}e^{-\sum\theta_{i}^{-1}y_{i}}
\]
Define the transform $\tilde{\varphi}:\,\mathbb{R}^{n}\rightarrow\mathbb{R}^{n}$
by: $\tilde{y}=\sum_{i}\theta_{i}^{-1}y_{i}$ and $\tilde{x}_{i}=\tilde{y}^{-1}\theta_{i}^{-1}y_{i}$
for $i\leq n-1$. This mapping is invertible and its Jacobian at $\tilde{y}$
is $\left(\prod_{i}\theta_{i}^{-1}\right)\tilde{y}$. Thus the joint
density of $\left(\tilde{y},\tilde{x}\right)$ is:
\[
g\left(\tilde{y},\tilde{x}\right)\propto\left(1-\sum_{i=1}^{n-1}\tilde{x}_{i}\right)^{{k}_{n}-1}\prod_{i=1}^{n-1}\tilde{x}_{i}^{{k}_{i}-1}\tilde{y}^{\sum_{i}{k}_{i}-1}e^{-\tilde{y}}
\]
This shows that the variables $\tilde{y}$ and $\tilde{x}$ are independent,
and that the distribution of $\tilde{x}$ is a Dirichlet distribution.

Define the transform $\varphi:\, y=\sum_{i}y_{i}$ and $x_{i}=y^{-1}y_{i}$
for $k\leq n-1$. This mapping is also invertible, with Jacobin $y^{k}$.
The joint density of $\left(y,x\right)$ is:
\[
g\left(y,x\right)\propto\left(1-\sum_{i=1}^{n-1}x_{i}\right)^{{k}_{n}-1}\prod_{i=1}^{n-1}x_{i}^{{k}_{i}-1}y^{\sum_{i}{k}_{i}-1}e^{-y\sum_{i}\theta^{-1}x_{i}}
\]
By identification, we get: $g\left(x|y=1\right)=g\left(\Delta^{-1}\tilde{x}\right)$
with $\Delta$ the diagonal matrix defined by $\Delta_{ii}=\theta_{i}$.
Since $\tilde{\varphi}^{-1}\left(\left(\Delta^{-1}\tilde{x},1\right)^{T}\right)=\tilde{x}$,
the result ensues.
\end{IEEEproof}
\textbf{Sampling from conditional gamma distributions.} Consider a
set of $n$ independent Gamma distributions $T_{i}\sim\Gamma\left(k_{i},\theta_{i}\right)$,
a $n$-dimensional vector of positive numbers $\alpha\in\left(\mathbb{R}_{+}^{*}\right)^{n}$
and $t>0$. The purpose of this section is to present some practical
formulas to sample and compute the density function of the conditional
distribution:

\[
Z=T\bigg|\sum_{i}\alpha_{i}T_{i}=t
\]

We define this distribution over the $n$-dimensional simplex 
\[
\mathcal{S}_{\alpha,t}=\left\{ x\in\left(\mathbb{R}^{+}\right)\bigg|\alpha^{T}x=t\right\} 
\]
As before, we define a new measure over the hyperplane defined by
$\alpha^{T}x=t$, and use it as our base measure $\text{d}z$ for
$Z$. We call $f$ the probability density function of variable $Z$
with respect to this measure. With respect to this measure, the probability
density function of $Z$ is that of a Gamma-Dirichlet distribution:
\[
f\left(z\right)=\frac{1}{n}t^{n-1}\frac{\sqrt{\sum_{i}\alpha_{i}^{2}}}{\prod_{i=1}^{n}\alpha_{i}}f_{\Gamma D}\left(y;k,\hat{\theta}\right)
\]

with:
\[
\hat{\theta}_{i}=t^{-1}\alpha_{i}\theta_{i}
\]

\begin{IEEEproof}
Call $h$ the scaling transform $x:\mathbb{R}^{n+1}\rightarrow\mathbb{R}^{n+1}$
defined by $y=h\left(x\right):y_{i}=\alpha_{i}t^{-1}x_{i}$. Then
all points from $\mathcal{S}_{\alpha,t}$ are mapped into the $n$-regular
simplex $\mathcal{S}^{n-1}$. Since this transform is a linear scaling,
one only needs to find the volume of $\mathcal{S}_{\alpha,t}$ to
define the new probability distribution. Call $\tilde{\mathcal{S}}_{\alpha,t}$
the $n$-standard simplex defined by the origin coordinate $\mathbf{0}$
and by its vertices $\left(t^{-1}\alpha_{i}\mathbf{e}_{i}\right)_{i\in\left[1,n\right]}$.
Its volume is $\left|\tilde{\mathcal{S}}_{\alpha,t}\right|=\frac{1}{n!}t^{-n}\prod_{i}\alpha_{i}$.
The volume of the hypersurface $\mathcal{S}_{\alpha,t}$ can be found
by differentiating the scaled volume $s\tilde{\mathcal{S}}_{\alpha,t}$
along the normal axis $\left\Vert \alpha\right\Vert _{2}^{-1}\alpha$
for $s=1$: 
\[
\left|\mathcal{S}_{\alpha,t}\right|^{-1}=\frac{\text{d}\left|\tilde{\mathcal{S}}_{\alpha,t}\right|}{\text{d}s}\bigg|_{s=1}=\frac{1}{\left(n-1\right)!}t^{n-1}\frac{\left\Vert \alpha\right\Vert _{2}}{\prod_{i=1}^{n}\alpha_{i}}
\]
Since the volume of the $n$-regular simplex $\mathcal{S}^{n-1}$
is $\frac{1}{n!}$, we get our result.
\end{IEEEproof}
Because of this equivalence between Gamma-Dirichlet distribution and
independent Gamma distributions conditioned on a hyperplane, we can
use the straightforward sampling algorithm from the definition of
the Gamma-Dirichlet distribution to sample values from the conditioned
Gamma distribution. This algorithm is presented in Algorithm \ref{alg:Sampler-for-Gamma}.

%
%

%% file: code.tex
\section*{Appendix B: code snippet for the Spark program and the Streaming Spark
program}

Main program, written in Scala using Spark Streaming.
\renewcommand{\ttdefault}{txtt}
\lstset{language=scala}
\begin{lstlisting}
// Creates the spark context
val ssc = new StreamingContext(...)

// Build the stream
val data_stream = createDStreamFromConfig(ssc, config)

// Asynchronous stream computations start here.

// Create samples
val em_complete_samples = data_stream.flatMap({
  case (p_ob, w) =>
    val state = DistributedMutableStorage.state
    val complete_samples = GammaSample.getSamples(
	p_ob,
	state,
	num_em_samples)
    complete_samples.map((_, w))
})

// Shuffling
// First double is tt, second double is weight
val samples_by_node = {
  em_complete_samples
    .flatMap({ case (cs, w) =>
      (cs.mask.nodes zip cs.partial).map({
        case (node, partial_tt) => 
	(node, (partial_tt, w))
      })
    })
    .groupByKey
}

val new_node_states = samples_by_node.map({ 
  case (nid, values) => 
  val tts = values.map(_._1).toArray
  val ws = values.map(_._2).toArray
  val new_state = GammaLearn.maxLikelihood(tts, ws)
  (nid, new_state)
})

new_node_states.foreachRDD(state_rdd => {
  // This code will be called on the master node
  // every time a new RDD is received.
  val new_state = state_rdd.collect.toMap
  println("Computed new state: "+new_state)
})
\end{lstlisting}